\documentclass[letterpaper]{article} 
\usepackage{aaai24}  
\usepackage{times}  
\usepackage{helvet}  
\usepackage{courier}  
\usepackage[hyphens]{url}  
\usepackage{graphicx} 
\def\UrlFont{\rm}  
\usepackage{natbib}  
\usepackage{caption} 
\usepackage{algorithm}
\usepackage{algorithmic}

\usepackage{multirow}
\usepackage{booktabs}
\usepackage{diagbox}
\usepackage{makecell}
\usepackage{etoolbox}

\usepackage{amsmath}
\usepackage{mathrsfs,mathtools, amssymb}
\usepackage{bm}

\usepackage{subfigure}
\usepackage{appendix}

\usepackage{newfloat}
\usepackage{listings}
\DeclareCaptionStyle{ruled}{labelfont=normalfont,labelsep=colon,strut=off} 
\floatstyle{ruled}
\newfloat{listing}{tb}{lst}{}
\floatname{listing}{Listing}
\usepackage{xspace}
\newcommand{\model}{A2GNN\xspace}
\newcommand{\propagate}{propagation\xspace}
\newcommand{\MLP}{transformation\xspace}
\newcommand{\generalizability}{generalization capability\xspace}

\newcommand{\Fig}{Figure\xspace}
\newcommand{\Tab}{Table\xspace}
\newcommand{\Fom}{Formula\xspace}

\title{Rethinking Propagation for Unsupervised Graph Domain Adaptation}
\author {
    Meihan Liu\textsuperscript{\rm 1},
    Zeyu Fang\textsuperscript{\rm 1},
    Zhen Zhang\textsuperscript{\rm 3},
    Ming Gu\textsuperscript{\rm 1},
    Sheng Zhou\textsuperscript{\rm 1,2}\thanks{Corresponding author},
    Xin Wang\textsuperscript{\rm 4},
    Jiajun Bu\textsuperscript{\rm 1}
}
\affiliations {
    \textsuperscript{\rm 1}Zhejiang Provincial Key Laboratory of Service Robot, College of Computer Science, Zhejiang University \\
    \textsuperscript{\rm 2}School of Software Technology, Zhejiang University\\
    \textsuperscript{\rm 3}Department of Computer Science, National University of Singapore\\
    \textsuperscript{\rm 4}Department of Computer Science and Technology, Tsinghua University\\
    \{lmh\_zju, fangzeyu, guming444, zhousheng\_zju, bjj\}@zju.edu.cn, \\ 
    zhen@nus.edu.sg,
    xin\_wang@tsinghua.edu.cn 
}

\begin{document}

\maketitle

\begin{abstract}
Unsupervised Graph Domain Adaptation (UGDA) aims to transfer knowledge from a labelled source graph to an unlabelled target graph in order to address the distribution shifts between graph domains. Previous works have primarily focused on aligning data from the source and target graph in the representation space learned by graph neural networks (GNNs). However, the inherent \generalizability of GNNs has been largely overlooked. Motivated by our empirical analysis, we reevaluate the role of GNNs in graph domain adaptation and uncover the pivotal role of the propagation process in GNNs for adapting to different graph domains. We provide a comprehensive theoretical analysis of UGDA and derive a generalization bound for multi-layer GNNs. By formulating GNN Lipschitz for k-layer GNNs, we show that the target risk bound can be tighter by removing propagation layers in source graph and stacking multiple propagation layers in target graph. Based on the empirical and theoretical analysis mentioned above, we propose a simple yet effective approach called \model for graph domain adaptation. Through extensive experiments on real-world datasets, we demonstrate the effectiveness of our proposed \model framework.

\end{abstract}

\section{Introduction}
The past decade has witnessed remarkable achievements of Graph Neural Networks (GNNs) \cite{GCN,Degradation22,yang2022graph} across a wide range of applications, from social network analysis \cite{xia2021deepis} and protein prediction \cite{li2021structure}, to traffic flow forecasting \cite{lu2022make}, etc. As the distribution shifts commonly exist among real-world graphs, it makes the majority of existing GNNs fail to generalize to new domains. Recently, Unsupervised Graph Domain Adaptation (UGDA) has emerged as a key focus in both industrial and academic communities. The goal of UGDA is to facilitate the knowledge transfer from a labelled source graph to an unlabelled target graph, which addresses the issue of distribution shifts, thus unlocking the full potential of graph neural networks for diverse applications.

Inspired by the success of domain adaptation in computer vision and natural language processing \cite{Ganin2015UnsupervisedDA}, most existing UGDA methods have primarily concentrated on aligning source and target graph distributions in the representation space generated by GNNs \cite{zhu2023explaining, StruRW}.
This includes learning domain-invariant representations with measurements like maximum mean discrepancy \cite{CDNE}, graph subtree discrepancy \cite{GRADE}, or performing domain discrimination \cite{ACDNE, ASN, AdaGCN, UDAGCN, SpecReg}.
However, the underlying \generalizability of GNNs has been largely overlooked.
Given that the representations learned by GNNs play a pivotal role in UGDA, we propose to revisit the step prior to alignment and pose a fundamental research question: \textbf{\emph{Does GNNs have an inherent capability to adapt to new domains?}}

\begin{figure}[t]
    \centering
    \includegraphics[width=\linewidth]{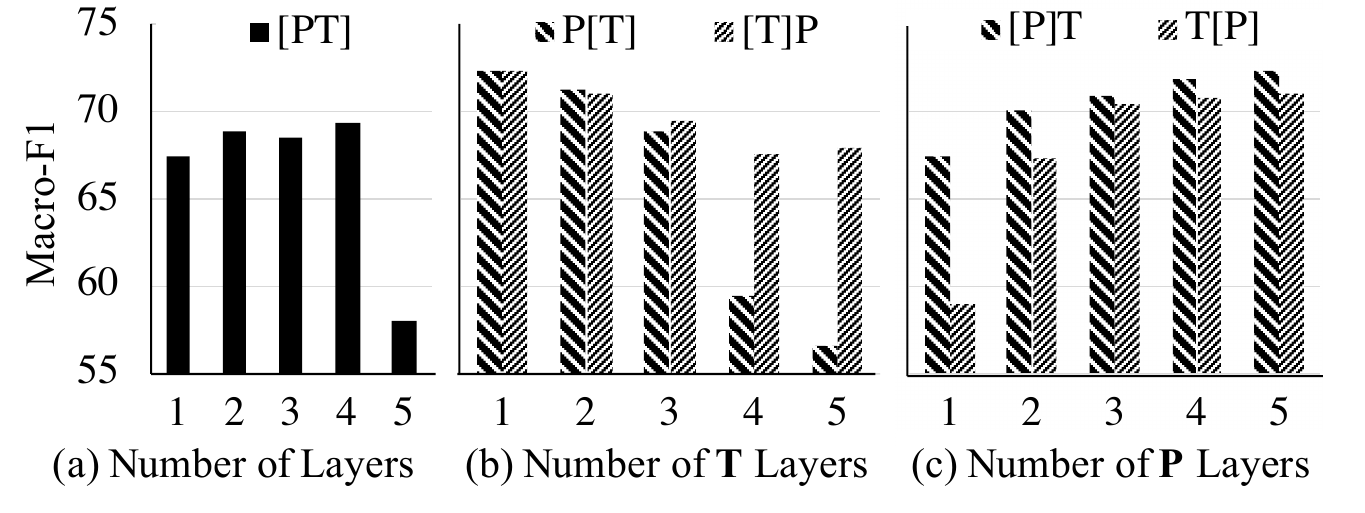}
    \caption{The influence of different operations in GNNs on task $D \rightarrow A$. $[\cdot]$ indicates the module stacking operation.}
    \label{fig:layers}
\end{figure}

To answer this question, we first decouple GNN's mechanism into two independent operations: {\it Propagation} (\textbf{P}) and {\it Transformation} (\textbf{T}), where the \textbf{P} operation performs message passing between neighboring nodes and the \textbf{T} operation conducts linear transformations to the node representations. Following the disentanglement, the GNN's architecture can be constructed with the combinations of \textbf{P} and \textbf{T}. \Fig \ref{fig:layers} presents the results under different architectures, from which we have the following key observations:
\textit{(i) Simultaneously adding P and T does not substantially enhance its \generalizability}. As we can see from \Fig \ref{fig:layers}(a), an incremental trend can be observed when the number of layers is fewer than 4, and more layers will result in decreased performance.
\textit{(ii) Solely adding T may impair its adaptation capability}. The UGDA performance in \Fig \ref{fig:layers}(b) shows a significant decrease with the addition of more transformation layers while keeping the number of propagation layers fixed; 
\textit{(iii) Operation P plays a crucial role in enhancing its adaptation capability}. Interestingly, results in \Fig \ref{fig:layers}(c) demonstrates that increasing propagation layers can achieve significant performance improvements compared with increasing both transformation and propagation layers.

Motivated by the above findings, we revisit the GNN's architecture in unsupervised graph domain adaptation and propose a simple yet effective model named \underline{\textbf{A}}symmetric \underline{\textbf{A}}daptive \underline{\textbf{GNN}} (abbreviated as A2GNN\footnote{https://github.com/Meihan-Liu/24AAAI-A2GNN}). While most existing UGDA approaches use a shared architecture between source and target graph, we argue that it's not necessary as we thought. By introducing a single transformation layer on source graph and stacking multiple propagation layers on target graph, we are able to achieve superior performance.
Furthermore, we also provide a comprehensive theoretical analysis for our proposed A2GNN via the graph filter theory and prove that the generalization gap of multi-layer GNNs depends on its \propagate layers. With the designed asymmetric architecture, our proposed A2GNN exhibits a tighter target risk bound. 
Empirical results on multiple real-world datasets demonstrate that \model outperforms recent state-of-the-art baselines on node classification tasks with different gains. 
We highlight our contributions as follows:
\begin{itemize}
    \item \textbf{New Perspective}. Our work is the first to investigate the underlying \generalizability of GNNs in UGDA task. Remarkably, our method outperforms state-of-the-art graph domain adaptation approaches.
    \item \textbf{Simple but Effective Method}. Based on the aforementioned observations, we propose an embarrassingly simple yet effective method \model with asymmetric network architecture. Comprehensive experimental results show the effectiveness of our proposed \model.
    \item \textbf{Theoretical Analysis}. In addition to investigating the underlying \generalizability of GNNs, we also derive a domain adaptation bound for multi-layer GNNs. Furthermore, we prove that our proposed \model yields a tighter error bound.
\end{itemize}

\section{Related Work}

Unsupervised domain adaptation (UDA) involves the transfer of knowledge from a labelled source domain to an unlabelled target domain \cite{10TransferLearningSurvey, 18DomainAdaptationSurvey}. A commonly used approach in domain adaptation is to minimize the domain discrepancy and learn domain-invariant representations \cite{Ganin2015UnsupervisedDA, Long2015LearningTF, Long2017ConditionalAD, Pei2018MultiAdversarialDA, Tzeng2017AdversarialDD}, where great success has been achieved in computer vision and natural language processing communities \cite{blitzer2006domain,venkateswara2017deep}. However, these methods assume that their inputs are independently and identically distributed data, thus they are not appropriate for tasks involving non-IID data, such as node classification in graph-structured data.

Recently, several approaches have been proposed to transfer knowledge across graph-structured non-IID data. These methods can be broadly categorized into two classes: minimizing pre-defined domain discrepancy metrics and adversarial learning techniques. 
For the pre-defined domain discrepancy metrics minimization models \cite{CDNE, GRADE}, node representations are first obtained from the encoder, and then domain-invariant representations are learned by minimizing pre-defined domain discrepancy metrics, such as MMD \cite{MMD}, graph subtree discrepancy \cite{GRADE}, etc.
Instead of explicitly minimizing domain discrepancies, some methods incorporate the encoder with a domain classifier that predicts the domain from which the representation originates. Among them, UDAGCN \cite{UDAGCN} and AdaGCN \cite{AdaGCN} integrates graph convolution with adversarial training for graph transfer learning. ACDNE \cite{ACDNE} utilizes Gradient Reversal Layer (GRL) \cite{GRL} to make node representation domain-invariant. ASN \cite{ASN} further improves node representations with attention mechanisms and disentanglement, where domain-private and domain-shared information are preserved. SpecReg \cite{SpecReg} derives the generalization bound for one-layer GNNs and proposes spectral regularization to restrict the bound. 
Instead of focusing on minimizing domain discrepancy, in this paper, we investigate the GNN's underlying \generalizability behind its architecture and propose a simple yet effective method for unsupervised graph domain adaptation.

\section{Preliminaries}

\subsubsection{Notations.}
Consider an attributed graph $G = (\mathcal{V}, \mathcal{E})$ with $n$ nodes and $m$ edges. We denote $\mathbf{X} = \{{x_v|v \in } \mathcal{V} \} \in \mathbb{R}^{n \times d}$ as node attribute matrix, where $d$ is the dimension of node attributes. The adjacency matrix is denoted as $\mathbf{A} \in \mathbb{R}^{n \times n}$, where $\mathbf{A}_{i, j}=1$ means there exists an edge $e_{i,j} \in \mathcal{E}$ connecting node $v_i$ and $v_j$, and $\mathbf{A}_{i, j}=0$ otherwise. For a node classification task, we denote the prediction targets by $\mathcal{Y} \in \mathbb{R}^{n \times C}$, where $C$ is the number of classes.

\subsubsection{Problem Definition.}
Unsupervised Graph Domain Adaptation (UGDA) aims to transfer the knowledge from a labelled source graph to an unlabelled target graph.
Formally, given graph $G_S = (\mathcal{V}_S, \mathcal{E}_S, \mathcal{Y}_S)$ and graph $G_T = (\mathcal{V}_T, \mathcal{E}_T)$ with the covariate shift assumption \cite{BenDavid2006, BenDavid2009ATO} that $\mathbb{P}_S(G) \neq \mathbb{P}_T(G)$ and $\mathbb{P}_{S}(Y|G)=\mathbb{P}_T(Y|G)$, where $\mathbb{P}_S$ and $\mathbb{P}_T$ are the probability distributions of the source and target domains, respectively. The target graph label $\mathcal{Y}^t$ is unknown. Our goal is to train a graph neural network (GNN) $h: \mathcal{G} \rightarrow \mathcal{Y}$ using the labeled source graph $G^s$ and the unlabeled target graph $G^t$ to accurately predict labels $\hat{y}^t$ on target graph $G^t$.

\section{Rethinking Propagation for UGDA}
In this section, we start with an empirical analysis of GNNs in the context of UGDA, focusing particularly on the propagation and transformation process. Guided by the intriguing insights obtained from our empirical analysis, we introduce a straightforward yet effective framework called \model for unsupervised graph domain adaptation. Lastly, we offer a theoretical analysis of the proposed approach.

\subsection{Empirical Analysis}
Although the Introduction presented several observations, this section provides more detailed elaboration and additional information. In general, most existing GNNs \cite{GCN,wu2019simplifying,klicpera2018predict} follow the neural message passing mechanism, which learns node representations by recursively aggregating information from neighborhood nodes. Specifically, the message passing procedure contains two major steps, namely {\it Transformation} (\textbf{T}) and {\it Propagation} (\textbf{P}) \cite{hamilton2017inductive,Degradation22}.
Let $\mathbf{H}^{(l-1)}$ denote the node representations generated by the previous layer and we set $\mathbf{H}^0 = \mathbf{X}$ at the beginning of the GNN. The \textbf{P} operation can be donated as $\textbf{P}(\mathbf{H}^{(l-1)}) = \hat{\mathbf{A}}\mathbf{H}^{(l-1)} $, where $\hat{\mathbf{A}}$ is the normalized adjacency matrix with self-connections. The \textbf{T} operation applies non-linear transformations to the node representations, which can be donated as $ \textbf{T}({\mathbf{H}^{(l)}}) = \sigma(\mathbf{H}^{(l)} \mathbf{W}^{(l)}) $, where $\sigma$ is the non-linear activation function and $\mathbf{W}$ is the learnable weight matrix.

To investigate the role of transformation and propagation operations in UGDA task, we use the classical metric MMD \cite{MMD} to minimize the domain discrepancy on task $D \rightarrow A$. 
The implementation details is provided in Appendix. According to the results, we have the following interesting and inspiring findings:

\begin{table}
    \small
    \centering
    \begin{tabular}{l|c|c|c}
    \toprule[0.8pt]
    Architecture & [\textbf{PT}] & [\textbf{P}]\textbf{T} & [\textbf{T}]\textbf{P} \\
    \midrule[0.8pt]
    Macro-F1    & 68.86    &72.41   & 70.61  \\
    \bottomrule[0.8pt]
    \end{tabular}
    \caption{Performance on different architectures.}
    \label{tab:obervation_architectures}
\end{table}

\begin{table}
\centering
\small
    \begin{tabular}{c|c|c}
    \toprule[0.8pt]
    \diagbox{target}{source} & w/o \propagate & w/ \propagate	\\
    \midrule[0.8pt]
         w/o \propagate  & 57.43       &55.68    \\
    \midrule[0.4pt]
         w/   \propagate  & \textbf{75.43} & 72.41  \\
    \bottomrule[0.8pt]
    \end{tabular}
    \caption{Macro-F1 with/without \propagate in graphs.}
    \label{tab:obervation_2}
\end{table}

\subsubsection{Observation 1: Propagation operation reflects the \generalizability of GNNs.}

Since we have decoupled message passing procedure into operations \textbf{T} and \textbf{P}, existing GNNs can be broadly categorized into the following three types: [\textbf{PT}] which means stacking multiple \textbf{P} and \textbf{T} operations simultaneously; [\textbf{P}]\textbf{T} which indicates stacking multiple propagation operations without changing \textbf{T}; [\textbf{T}]\textbf{P} which represents stacking multiple transformation operations without changing \textbf{P}; [$\cdot$] is the stacking operation. 
We first search the optimal architecture for each category and report their mean performance in Table \ref{tab:obervation_architectures}. As we can see, both [\textbf{P}]\textbf{T} and [\textbf{T}]\textbf{P} architectures outperform the [\textbf{PT}] architecture, indicating that the network architecture in UGDA indeed affects its performance. To further analyze the advantage of the [\textbf{P}]\textbf{T} and [\textbf{T}]\textbf{P} architectures, we evaluate its effects by increasing the number of \textbf{P} or \textbf{T} operations. The results, already depicted in Figure \ref{fig:layers}, demonstrate that stacking \textbf{P} operations improves its performance in both architectures, while stacking \textbf{T} operations has the opposite effect. 
Based on these findings, we conclude that (\textit{i}) \textbf{P} \textit{operation plays a pivotal role in determining the generalization ability of the model}, and (\textit{ii}) \textit{The} \textbf{T} \textit{operations are only needed to be executed once}.

\subsubsection{Observation 2: Conducting \propagate solely on target graph is of primary importance.} 

Although we have observed that \propagate contributes to the improvement of \generalizability in GNNs, the impacts of \propagate on both graphs remain unclear. 
To investigate this issue, we use the [\textbf{P}]\textbf{T} architecture and examine the effect of \propagate on different domains while keeping the number of \textbf{T} operations unchanged. When comparing the second and first rows of Table \ref{tab:obervation_2}, we find \propagate on the target graph is the primary factor that affects the model's \generalizability, while conducting \propagate solely on the source graph produces the worst performance. Interestingly, performing \propagate solely on the target graph leads to better performance compared to performing it on both graphs (yielding a 1.85\% increase in Macro-F1). These observations support the idea that \propagate has different effects on different domains. Specifically,  (\textit{iii}) \textit{stacking} \textbf{P}  \textit{operations on the target graph is particularly important}, and  (\textit{iv}) \textbf{P}  \textit{operations are not necessary on the source graph}.

\subsection{The Proposed Framework}

\begin{figure}[t]
    \centering
    \includegraphics[width=0.92\linewidth]{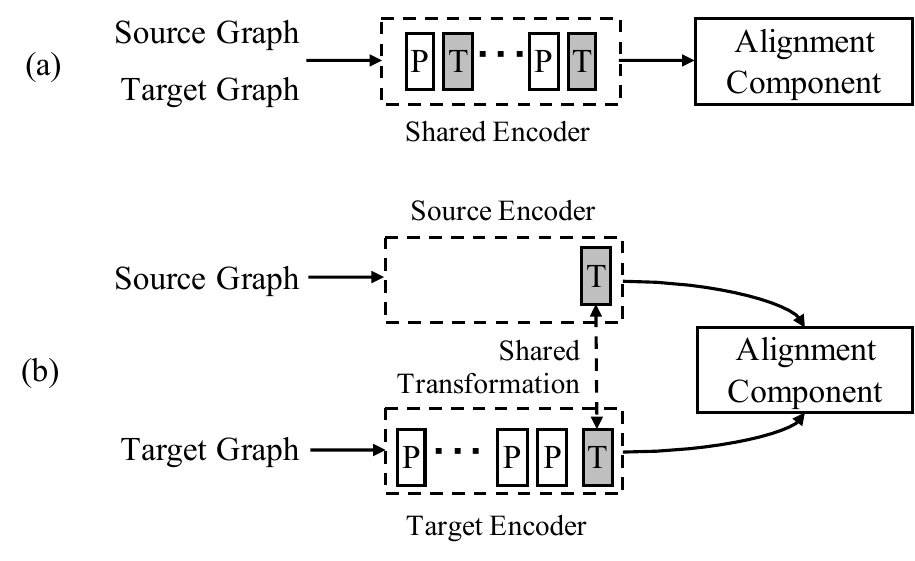}
    \caption{The framework of existing models and our proposed A2GNN model.}
    \label{fig:framework}
\end{figure}

The aforementioned two observations unveil the inherent \generalizability of GNNs. Building upon these insights, we propose an asymmetric graph domain adaptation model called \model, which is embarrassingly simple yet extremely effective. The overall framework is illustrated in \Fig \ref{fig:framework}(b), while \Fig \ref{fig:framework}(a) demonstrates the symmetric architecture commonly employed by existing methods \cite{ACDNE,AdaGCN}. 

According to the analyses in the previous section, we find that it's not necessary to utilize a shared GNN architecture between source and target graph, meanwhile stacking more propagation layers and using fewer transformation layers on target graph could lead to better performance. Therefore, we initialize our \model with the following asymmetric architecture.  
For the source graph branch, we remove all the propagation layers, then it will degenerate to a non-linear MLP, i.e., $\mathbf{H}^s=\sigma\left(\mathbf{X} \mathbf{W}\right)$, where $\mathbf{X}$ is the node feature matrix and $\mathbf{W}$ denotes the learnable weight matrix. 
While for the target graph branch, we stack $k$ \propagate layers with one transformation layer, i.e., $\mathbf{H}^t=\sigma\left(\mathbf{A}^k \mathbf{X} \mathbf{W}\right)$. 
Note that the learnable weight matrix $\mathbf{W}$ is shared across the source and target graph to prevent significant divergence between their representations.

The objective function of the unsupervised graph domain adaptation framework can be formulated as follows:
\begin{equation}
    \mathcal{L}_{GDA} = \mathcal{L}_{cls} + \alpha \cdot \mathcal{L}_{align},
    \label{overall-loss}
\end{equation}
where the first term $\mathcal{L}_{cls}$ indicates the cross-entropy loss function for the node classification task using source graph labels, while the second term $\mathcal{L}_{align}$ focuses on minimizing domain discrepancy to obtain domain-invariant representations. $\alpha$ serves as a trade-off parameter that balances the contributions of the two loss terms.

It is worth highlighting that our domain adaptation framework is flexible to incorporate various alignment loss functions. In our experiments, we utilize two classical alignment components: maximum mean discrepancy (MMD) \cite{MMD} for explicit alignment, represented as $\mathcal{L}_{GDA} = \mathcal{L}_{cls} + \alpha \cdot \mathcal{L}_{mmd}$, and adversarial training mechanism \cite{goodfellow2020generative} with gradient reversal layer (GRL) \cite{GRL} for implicit alignment, represented as $\mathcal{L}_{GDA} = \mathcal{L}_{cls} + \alpha \cdot \mathcal{L}_{adv}$. Furthermore, our domain adaptation framework offers the versatility to integrate various propagation mechanisms, and we will delve into the effects of different propagation mechanisms in the ablation study section.

\subsection{Theoretical Analysis}
In this subsection, we aim to provide a theoretical explanation for the efficacy of our proposed \model. Specifically, we first revisit the graph domain adaptation error bound formulated by \cite{SpecReg}, then we prove that the operation of \MLP is independent of tightening the error bound. Lastly, we show how our proposed \model narrows down the error bound.

\subsubsection{Theorem 1.}
Suppose the feature extractor $f$ and the classifier $c$ is $K_f$-Lipschitz and $K_c$-Lipschitz continuous, where the Lipschitz norm $\|f\|_{\text {Lip }}=\max _{G_1, G_2} \frac{\left\|f\left(G_1\right)-f\left(G_2\right)\right\|_2}{\eta\left(G_1, G_2\right)}=K_f$ holds for some graph distance measure $\eta$. Let $\mathcal{H}:=\{h: \mathcal{G} \rightarrow \mathcal{Y}\}$ be the set of bounded real-valued functions, i.e. $h=c \circ f \in \mathcal{H}$, with probability at least $1-\delta$ the target risk $\epsilon_{\mathrm{T}}(g, \hat{g})$ is bounded as in the following inequality \cite{SpecReg}:
\begin{equation}
\begin{aligned}
    \epsilon_{\mathrm{T}}(h, \hat{h}) 
    \leq \ 
    & \hat{\epsilon}_{\mathrm{S}}(h, \hat{h}) 
      + 2 K_f K_c D\left(\mathbb{P}_{\mathrm{S}}(G), \mathbb{P}_{\mathrm{T}}(G)\right)\\
    & + \sqrt{\frac{4 d}{n^s} \log \left(\frac{e n^s}{d}\right) + \frac{1}{n^s} \log \left(\frac{1}{\delta}\right)} 
    + \omega,
\end{aligned} 
\label{therorem_1}
\end{equation}
where $\hat{\epsilon}_{\mathrm{S}}(h, \hat{h})$ and $\hat{\epsilon}_{\mathrm{T}}(h, \hat{h})$ represent the empirical source and target risks respectively. $\hat{h}$ denotes the labelling function. $\omega=\min _{\|c\|_{\operatorname{Lip}} \leq K_c,\|f\|_{\operatorname{Lip}} \leq K_f}\left\{\epsilon_{\mathrm{S}}(h, h^*)+\epsilon_{\mathrm{T}}(h, h^*)\right\}$ is the combined error of the ideal hypothesis $h^*$ that we expect to be small. $D(\cdot)$ indicates the distance metric, which formulates the source and target error functions in a Reproducing Kernel Hilbert Space.

Theorem 1 indicates that the generalization gap depends on the domain divergence $2 K_f K_c D\left(\mathbb{P}_{\mathrm{S}}(G), \mathbb{P}_{\mathrm{T}}(G)\right)$, where the Lipschitz constant $K_f$ is a conceptual property related to the model that needs to be instantiated \cite{SpecReg}. Therefore, one way to tighten the bound is to regularize the Lipschitz constant $K_f$ of the feature extractor $f$.

Next, we construct a GNN by composing a graph filter and nonlinear mapping that $f\left(G\right)
=r\left(\sigma\left(\mathcal{S}\left(A\right) X W\right)\right)$, where $r$ is the mean/sum/max readout function to pool node representations. $\mathcal{S}$ is the polynomial function that $\mathcal{S}\left(A\right)=\sum_{k=0}^{\infty} s_k A^k$. $W \in \mathcal{R}^{D \times D^{\prime}}$ denotes the learnable weight matrix. The pointwise nonlinearity holds as $|\sigma(b)-\sigma(a)| \leq|b-a|, \forall a, b \in \mathcal{R}$. Based on these notations, a $l$-layer GNN can be constructed as $f(G) = r \circ f^{(l)} \circ f^{(l-1)}\circ \cdots \circ f^{(1)}(G)$. The $l$-layer GNN Lipschitz constant $K_f$ is instantiated in the following lemma.

\subsubsection{Lemma 1.}
Suppose that the edge perturbation is bounded that $\forall G_1, G_2 \in \mathcal{G}$, $EP = \left\|A_1-P^* A_2 P^{* \mathrm{~T}}\right\|_{\mathrm{F}} \leq \varepsilon$ with the optimal permutation $P^*$, and there exists an eigenvalue $\lambda^* \in \mathcal{R}$ to achieve the maximum $\left|\mathcal{S}\left(\lambda^*\right)\right|<\infty$. 
Assuming $\|X\|_{\mathrm{op}} \leq 1$ and $\|W\|_{\mathrm{op}} \leq 1$ ( $\|\cdot\|_{\mathrm{op}}$ stands for operator norm), we can then calculate the Lipschitz constant $l$-layer GNN as follows:
\begin{equation}
    \begin{aligned}
        K_f =\max \left\{
T_1^{(l)} + \sum_{i=1}^{l-1}{(\prod \limits_{j=i+1}^{l}T_2^{(j)})T_1^{(i)}}
, \prod \limits_{i=0}^l T_2^{(i)} \right\},
    \end{aligned}
    \label{lamma1}
\end{equation}
where $T_1^{(l)} = K_\lambda\left(1+\tau \sqrt{n}\right)EP+\varepsilon \cdot \mathcal{O}\left(EP^2\right)$ 
and $T^{(l)}_2 = \left|\mathcal{S}^{(l)}\left(\lambda_T\right)\right|$. $\tau$ stands for the eigenvector misalignment that can be bounded. 
$n=n_S + n_T$ means the total number of nodes in source and target graph. 
$\mathcal{O}\left( \cdot \right)$ is the remainder term defined in \cite{Gama2019Stability}, 
and $K_\lambda$ is the spectral Lipschitz constant that $\forall \lambda_i, \lambda_j,\left|\mathcal{S}\left(\lambda_i\right)-\mathcal{S}\left(\lambda_j\right)\right| \leq K_\lambda\left|\lambda_i-\lambda_j\right|$.

By integrating Theorem 1 and Lemma 1, we obtain the graph domain adaptation bound for a $l$-layer GNN. Lemma 1 indicates that (\textit{i}) \textit{decreasing the Lipschitz constant $K_f$ can tighten the target error bound in Theorem 1}, and (\textit{ii}) \textit{the \MLP operation is independent of the GNN Lipschitz constant} $K_f$. Therefore, it is reasonable to employ a single \MLP layer in conjunction with multiple \propagate layers in our proposed \model, which aligns with the empirical experiments in \textbf{Observation 1}. Based on the above theory, we next prove \model has a tighter target error bound.

\subsubsection{Lemma 2.}
Following the setting in Lemma 1, we use a single linear \MLP in conjunction with $k$ \propagate layers in \model as $f_M^{\left( l \right)}\left(G\right)=r\left(\sigma\left(\mathcal{S}^{\left( l \right)}\left(A^k\right) X W\right)\right)$. We rephrase $T_2$  as $T_2(x)$, $x$ is the number of \propagate layers. For $\forall k > 1$, there exists: 
\begin{equation}
    \begin{aligned}
    \prod \limits_{i=0}^l T_2^{(i)}(k) < \prod \limits_{i=0}^l T_2^{(i)}(1), 
    \end{aligned}
    \label{lemma2}
\end{equation}
Based on Lemma 2, we state that the second term of $K_f$ can be smaller when we have $k \ (k>1$) \propagate layers. Next, we show that the first term of $K_f$ can be smaller too.

\subsubsection{Lemma 3.} 
Following the setting in Lemma 2, if the source graph structure does not participate in the training procedure, then the corresponding GNN degenerates into $f_L\left(G\right)
=r\left(\sigma\left( X W\right)\right)$. We rephrase $T_1^{(l)}$ as $T_1^{(l)}(x)$, and $x$ indicates the number of \propagate layers on source graph. For $\forall k > 1$, there exists: 
\begin{equation}
    \begin{aligned}
     \resizebox{\hsize}{!}{$
     T_1^{(l)}(0) + \sum_{i=1}^{l-1}{(\prod \limits_{j=i+1}^{l}T_2^{(j)}(k))T_1^{(i)}(0)} 
     < T_1^{(l)}(1) + \sum_{i=1}^{l-1}{(\prod \limits_{j=i+1}^{l}T_2^{(j)}(1))T_1^{(i)}(1)}.
     $}
    \end{aligned}
    \label{lamma3}
\end{equation}
According to Lemma 3, we state that the first term of $K_f$ can be smaller when we conduct $k \ (k>1$) \propagate layers on the target graph and only one \MLP layer on the source graph. We denote the Lipschitz constant $K_f'$ of \model as follows:
\begin{equation}
    \begin{aligned}
     \resizebox{\hsize}{!}{$
    K_f' =\max \left\{
        T_1^{(l)}(0) + \sum_{i=1}^{l-1}{(\prod \limits_{j=i+1}^{l}T_2^{(j)}(k))T_1^{(i)}(0)}
        , \prod \limits_{i=0}^l T_2^{(i)}(k)
        \right\}.
     $}
    \end{aligned}
    \label{lamma4}
\end{equation}
With Lemma 2 and Lemma 3, we derive that $K_f' < K_f$. In other words, the Lipschitz constant $K_f'$ of \model is smaller than that of the previous frameworks. This conclusion is in line with \textbf{Observation 2}. Combining this conclusion with Theorem 1, we prove that the graph domain error bound of \model is tighter. Note that there is currently no theoretical evidence to support that more layers lead to tighter bound in \Fom (\ref{lamma4}). Investigation on $k$ is shown in \Fig \ref{fig:ablation}(a). A more detailed proof process is provided in the Appendix.

\begin{table}[]
\centering
\small
\begin{tabular}{c|c|c|c|c|c}
\toprule[0.8pt]
Types                      & Datasets        & \#Node  & \#Edge    & \#Feat           & \#Label              \\ 
\midrule[0.8pt]
\multirow{3}{*}{Citation} & ACMv9      & 9,360 & 15,556  & \multirow{3}{*}{6,775} & \multirow{3}{*}{5} \\
                          & Citationv1 & 8,935 & 15,098  &                        &                    \\
                          & DBLPv7     & 5,484 & 8,117   &                        &                    \\ 
\midrule[0.8pt]
\multirow{2}{*}{Social}   & Germany    & 9,498 & 153,138 & \multirow{2}{*}{3,170} & \multirow{2}{*}{2} \\
                          & England    & 7,126 & 35,324  &                        &                    \\ 
\bottomrule[0.8pt]
\end{tabular}
\caption{Dataset Statistics.}
\label{tab:datasets}
\end{table}

\begin{table*}[htbp]
    \centering
    \small 
    \begin{tabular}{l|cc|cc|cc|cc|cc|cc}
    \toprule[0.8pt]
    \multirow{2}{*}{Models} 	&  \multicolumn{2}{c|}{D $\rightarrow$ A} & \multicolumn{2}{c|}{C $\rightarrow$ A} &\multicolumn{2}{c|}{A $\rightarrow$ D}	& \multicolumn{2}{c|}{C $\rightarrow$ D}	& \multicolumn{2}{c|}{A $\rightarrow$ C} & \multicolumn{2}{c}{D $\rightarrow$ C} \\
    \cmidrule[0.8pt]{2-13}
                & Ma-F1 & Mi-F1 & Ma-F1 & Mi-F1 & Ma-F1 & Mi-F1 & Ma-F1 & Mi-F1 & Ma-F1 & Mi-F1 & Ma-F1 & Mi-F1	\\
    \cmidrule[0.8pt]{1-13}
    DeepWalk    & 19.83 & 26.23 & 19.33 & 21.94 & 19.87 & 25.94 & 17.51 & 22.57 & 17.72 & 21.05 & 22.76 & 29.46 \\ 
    node2vec    & 22.05 & 28.61 & 17.99 & 21.76 & 19.50 & 24.54 & 24.98 & 28.95 & 25.84 & 29.89 & 16.22 & 21.16 \\
    ANRL        & 19.12 & 29.56 & 22.04 & 31.84 & 23.33 & 29.54 & 22.71 & 25.90 & 20.93 & 30.31 & 18.25 & 25.99 \\
    \cmidrule[0.8pt]{1-13}
    GAT         & 43.95 & 52.93 & 42.14 & 50.37 & 41.36 & 53.80 & 45.25 & 55.85 & 43.64 & 57.13 & 50.04 & 55.52 \\
    GraphSAGE   & 57.31 & 59.22 & 64.69 & 65.22 & 61.80 & 64.82 & 66.86 & 69.96 & 69.14 & 71.40 & 64.90 & 67.85 \\
    GIN         & 56.50 & 58.98 & 59.48 & 60.46 & 50.49 & 59.10 & 63.48 & 66.27 & 62.49 & 68.61 & 63.21 & 69.25 \\
    GCN         & 59.42 & 63.35 & 70.39 & 70.58 & 65.29 & 69.05 & 71.37 & 74.53 & 74.78 & 77.38 & 69.79 & 74.17 \\ 
    \cmidrule[0.8pt]{1-13}
    CDNE        & 70.45 & 69.62 & $\underline{75.06}$ & $\underline{74.22}$ & 69.24 & 71.58 & 71.34 & 74.36 & 76.83 & 78.76 & 77.36 & 78.88 \\ 
    UDAGCN      & 55.89 & 58.16 & 67.22 & 66.80 & 64.83 & 66.95 & 69.46 & 71.77 & 60.33 & 72.15 & 61.12 & 73.28 \\
    ACDNE       & $\underline{72.64}$ & $\underline{71.29}$ & 74.79 & 73.59 & 73.59 & $\underline{76.24}$ & $\underline{75.74}$ & $\underline{77.21}$ & $\underline{80.09}$ & $\underline{81.75}$ & $\underline{78.83}$ & $\underline{80.14}$ \\
    ASN         & 71.49 & 70.15 & 73.17 & 72.74 & 71.40 & 73.80 & 73.98 & 76.36 & 77.81 & 80.64 & 75.17 & 78.23 \\
    AdaGCN      & 69.47 & 69.67 & 70.77 & 71.67 & 71.39 & 75.04 & 72.34 & 75.59 & 76.51 & 79.32 & 74.22 & 78.20 \\ 
    GRADE       & 59.35 & 63.72 & 69.34 & 69.55 & 63.03 & 68.22 & 70.02 & 73.95 & 72.52 & 76.04 & 69.32 & 74.32 \\
    SpecReg     & 72.34 & 71.01 & 73.15 & 72.04 & \underline{73.98} & 75.93 & 73.64 & 75.74 & 78.83 & 80.55 & 77.78 & 79.04 \\ 
    \cmidrule[0.8pt]{1-13}
    $\textbf{\model}_{adv}$ & \textbf{73.81} & \textbf{72.17} & \textbf{76.64} & \textbf{75.22} & \textbf{74.03} & 75.42 & \textbf{75.82} & \textbf{77.32} & 78.60 & 80.27 & 78.18 & 80.08 \\
    Improv.(\%)     & +1.17  & +0.88 & +1.58 & +1.00 & +0.05 & -0.82 & +0.08 & +0.11 & -1.49 & -1.48 & -0.65 & -0.06 \\
       $\textbf{\model}_{mmd}$ & $\textbf{75.69}$ & $\textbf{74.12}$ & $\textbf{77.57}$ & $\textbf{76.15}$ & $\textbf{75.78}$ & $\textbf{77.43}$ & $\textbf{77.04}$ & $\textbf{78.13}$ & $\textbf{81.30}$ & $\textbf{82.64}$ & $\textbf{79.74}$ & $\textbf{81.54}$ \\ 
    Improv.(\%)     & +3.05  & +2.83 & +2.52 & +1.93 & +1.80 & +1.19 & +1.30 & +0.92 & +1.21 & +0.89 & +0.91 & +1.40 \\
    \bottomrule[0.8pt]
    \end{tabular}
    \caption{Unsupervised node classification on citation network. The best result is bold and the second best is underlined.}
    \label{tab:main-results-citation}
\end{table*}

\section{Experiments and Analyses}

\subsection{Datasets} 
We conduct comprehensive experiments on three public citation networks and two social networks across a range of settings. These datasets are acquired from diverse sources and time periods with explicit covariate shifts. The citation networks\footnote{https://github.com/yuntaodu/ASN/tree/main/data} consist of three datasets: ACMv9, Citationv1, and DBLPv7. Each node represents a paper, and the edges indicate the citation relationships among them. \textbf{ACMv9 (A)} contains papers extracted from ACM between 2000 and 2010, \textbf{Citationv1 (C)} comprises papers obtained from Microsoft Academic Graph prior to 2008, and \textbf{DBLPv7 (D)} encompasses papers collected from DBLP during the period from 2004 to 2008. Our objective is to classify all the papers into five distinct research topics, i.e., Databases, Artificial Intelligence, Computer Vision, Information Security, and Networking. As for social networks, we choose Twitch gamer networks\footnote{http://snap.stanford.edu/data/twitch-social-networks.html}, which are collected from different regions. Each node within these networks represents a user, and the connections between them indicate their friendships. We extract node features encompassing information about users' gaming preferences, geographical locations, streaming habits, etc. Specifically, we focus on the two largest graphs: \textbf{Germany (DE)} and \textbf{England (EN)}. In this scenario, users are classified into two groups based on whether they employ explicit language. For a comprehensive overview of these datasets, please refer to \Tab \ref{tab:datasets}.

\subsection{Baselines} 
We compare our proposed \model with three categories of baselines:
(\textit{i}) {\it Hypothesis transfer with unsupervised graph representation learning}: DeepWalk \cite{DeepWalk}, node2vec \cite{node2vec}, and ANRL \cite{ANRL}. These methods first learn node representations in an unsupervised manner, and then the target graph node representations are evaluated using a classifier that has been trained on the source graph.
(\textit{ii}) {\it Source-only graph neural networks}: GCN \cite{GCN}, GAT \cite{GAT}, GraphSAGE \cite{GraphSAGE}, and GIN \cite{xu2018powerful}. This group of methods trains graph neural networks on the source graph in an end-to-end manner, thus they can be directly applied to target graph for evaluation.
(\textit{iii}) {\it Graph domain adaptation methods}: CDNE \cite{CDNE}, AdaGCN \cite{AdaGCN}, ACDNE \cite{ACDNE}, UDAGCN \cite{UDAGCN}, ASN \cite{ASN}, GRADE \cite{GRADE}, and SpecReg \cite{SpecReg}. Compared with the previous two groups, approaches in this group are specifically designed to address the graph domain adaptation problem, which are competitive baselines.

\subsection{Parameter Settings} 
Following previous works \cite{UDAGCN,ACDNE}, we use all the source nodes and target nodes for model training. Among them, 80\% of the labelled source nodes are utilized to provide supervision signals. The remaining labelled source nodes compose the validation set, and the evaluation is carried out on the target nodes. To ensure a fair comparison, we utilize the source codes released by the authors for each baseline, and then fine-tune their hyper-parameters to their optimal values. The node representation dimension is uniformly set as 128 for all the methods. Our proposed \model is implemented with PyTorch \cite{PyTorch}, where the learning rate and weight decay are searched in the range of $\{1e^{-1}, 1e^{-2}, 1e^{-3}, 1e^{-4}, 5e^{-4}\}$. We consider two variants of our proposed model: ${\rm \model}_{mmd}$ which uses Gaussian kernel as the selected kernel for MMD \cite{CDNE} and ${\rm \model}_{adv}$ which employs adversarial training for invariant representation learning \cite{ACDNE}. The experiments are repeated five times, and we report the mean performance in terms of Micro-F1 and Macro-F1 scores.

\subsection{Main Results}
Table \ref{tab:main-results-citation} shows the node classification performance on citation network, and we have the following observations: (\textit{i}) Our proposed \model outperforms recent state-of-the-art baselines with different gains. On average, ${\rm \model}_{mmd}$ achieves 1.80\% improvement in Macro-F1 and 1.53\% improvement in Micro-F1 scores under various settings, while ${\rm \model}_{adv}$ also yields comparable results. (\textit{ii}) The graph domain adaptation methods outperform the vanilla graph neural networks and the hypothesis transfer approaches, indicating that considering the domain discrepancy across networks is crucial. 
Furthermore, it should be noted that ${\rm \model}_{mmd}$ surpasses ${\rm \model}_{adv}$ by varying margins. This divergence can be attributed to the fact that MMD offers explicit signals, whereas adversarial training solely provides implicit signals.

Similarly, we can draw similar conclusions when it comes to social networks. The results are shown in Table \ref{tab:class-social}. In general, our proposed \model outperforms or shows comparable performance among all the settings. More specifically, ${\rm \model}_{adv}$ outperforms all the baselines on all tasks. ${\rm \model}_{mmd}$ obtains the highest scores on both Macro-F1 and Micro-F1 for the $EN \rightarrow DE$ task, and it ranks second in terms of Micro-F1 for the $DE \rightarrow EN$ task. On average, ${\rm \model}_{mmd}$ achieves 1.88\% improvement in Macro-F1 and 1.13\% improvement in Micro-F1, while ${\rm \model}_{adv}$ achieves a 2.54\% increase in Macro-F1 and 1.93\% improvement in Micro-F1. 
Moreover, we also observe that some domain adaptation methods are outperformed by source-only graph neural networks, which is known as negative transfer. This can be attributed to the heterophilic nature of these social graphs, and it introduces significant challenges for domain adaptation tasks. 
In conclusion, our proposed simple yet effective \model consistently exhibits impressive performance on different types of graphs, thus verifying its capability in graph domain adaptation task.

\subsection{Ablation Study}

\subsubsection{The impact of the number of \propagate layers $k$.}
\label{sec:ablation_propatation}

\begin{figure}[t]
    \centering
    \includegraphics[width=0.9\linewidth]{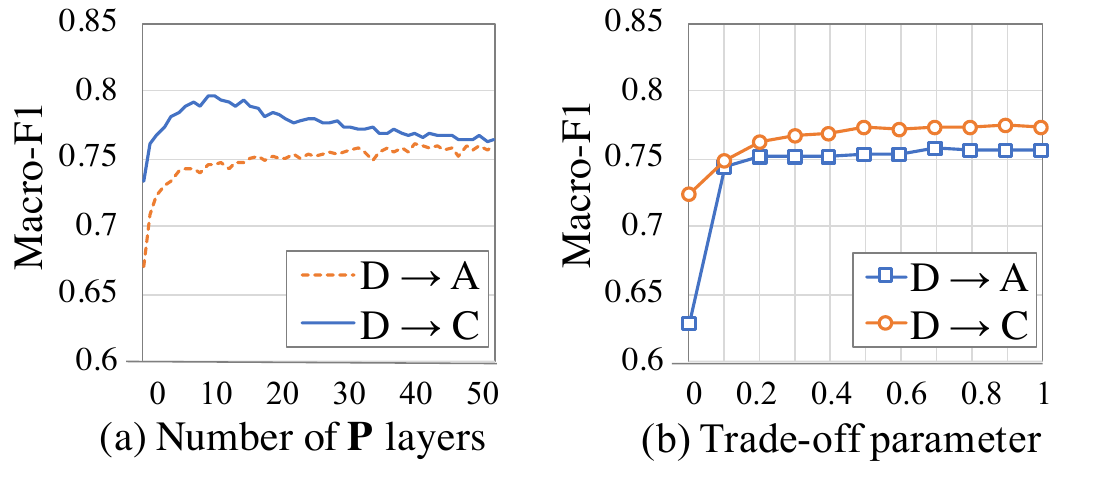}
    \caption{(a) shows the sensitivity of the num of \propagate layers $k$. (b) shows the sensitivity of trade-off parameter $\alpha$.}
    \label{fig:ablation}
\end{figure}

As we discussed in previous sections, Lemma 2 and Lemma 3 have proved that \propagate on the target graph for $k \ (k>1)$ times can tighten the target error bound. Therefore, we conducted some experiments to analyze the sensitivity of $k$, which is shown in \Fig \ref{fig:ablation}(a). On the one hand, utilizing a small number of $k \ (k<5)$ can significantly improve its performance. On the other hand, varying values of $k$ yield distinct effects across various tasks. For the $D \rightarrow A$ task, as the value of $k$ increases, its performance improves correspondingly. In contrast, for the $D \rightarrow C$ task, optimal performance is observed within the range of 5 to 15, followed by a gradual decline when $k$ surpasses 15. This decline is particularly pronounced in terms of Macro-F1 scores. In conclusion, the empirical results align with our theoretical analyses.

\begin{table}[htbp]
    \small
    \centering
    \begin{tabular}{l|cc|cc}
    \toprule[0.8pt]
    \multirow{2}{*}{Models}	& \multicolumn{2}{c|}{DE$\rightarrow$EN} & \multicolumn{2}{c}{EN$\rightarrow$DE}	\\
            \cmidrule[0.8pt]{2-5}
                 & Ma-F1 & Mi-F1 & Ma-F1 & Mi-F1 \\
            \cmidrule[0.8pt]{1-5}
        Node2vec & 46.96 & 52.64 & 50.10 & 54.61 \\
        DeepWalk & 46.54 & 52.18 & 49.97 & 55.08 \\
            \cmidrule[0.8pt]{1-5}
        GAT      & 49.50 & 54.84 & 40.08 & 43.65 \\
        GCN      & 54.55 & 54.77 & 51.04 & 62.03 \\
        GIN      & 49.91 & 52.39 & 44.26 & 55.26 \\
            \cmidrule[0.8pt]{1-5}
        UDAGCN    & \underline{58.19} & \underline{59.74} & 56.35 & 58.69 \\
        ACDNE    & 56.31 & 58.08 & \underline{57.92} & 58.79 \\
        ASN      & 51.21 & 55.45 & 45.90 & 60.45 \\
        AdaGCN   & 35.30 & 54.56 & 31.18 & 40.22 \\
        GRADE    & 56.38 & 56.40 & 56.83 & 61.18 \\
        SpecReg  & 50.30 & 56.43 & 46.13 & \underline{61.45} \\ 
            \cmidrule[0.8pt]{1-5}
        $\textbf{\model}_{mmd}$ & \textbf{58.44} & 59.54 & \textbf{61.42} & \textbf{63.90} \\
        Improv.(\%)     & +0.25  & -0.20 & +3.50 & +2.45 \\
        $\textbf{\model}_{adv}$ & \textbf{59.16} & \textbf{59.94} & \textbf{62.03} & \textbf{65.11} \\
        Improv.(\%)     & +0.97  & +0.20 & +4.11 & +3.66  \\
            \bottomrule[0.8pt]
    \end{tabular}
    \caption{Unsupervised node classification on social network. The best result is bold and the second best is underlined.}
    \label{tab:class-social}
\end{table}

\subsubsection{The influence of different \propagate schemes.}
To further investigate whether our proposed framework can be employed to various types of propagation schemes, we replace the widely used symmetric normalized transition matrix $\mathbf{P}_{\text {sym }}=\tilde{\mathbf{D}}^{-\frac{1}{2}} \tilde{\mathbf{A}} \tilde{\mathbf{D}}^{-\frac{1}{2}}$ in GCN with three commonly used transition matrices \cite{hassani2020contrastive}: 
(1) $\mathbf{P}_{\text {no-loop }}=\mathbf{D}^{-\frac{1}{2}} \mathbf{A} \mathbf{D}^{-\frac{1}{2}}$, i.e., removing the self-loops; 
(2) $\mathbf{P}_{\mathrm{rw}}=\tilde{\mathbf{D}}^{-1} \tilde{\mathbf{A}}$, i.e., the random walk matrix; 
(3) $\mathbf{P}_{\text {diff }}=\sum_{p=0}^{\infty} \frac{1}{e \cdot p !}\left(\tilde{\mathbf{D}}^{-1} \tilde{\mathbf{A}}\right)^p$, i.e., the heat kernel diffusion matrix, where $p$ is set as 10 by default. The results are presented in \Tab \ref{tab:schemes}. We observe that the relative performance stays at almost the same level when removing self-loop, which verifies our observation that deep neighbor information is more important than local information. The performance of random walk matrix drops on $D \rightarrow C$ task while staying at almost the same level on $D \rightarrow A$ task. The possible reason is that random walk matrix have incomplete neighbor information, which affects the model's ability to capture the structural information and thus affects its \generalizability. We also note a relatively deteriorated performance achieved by the diffusion transition matrix. This can be attributed to its emphasis on global information, neglecting the crucial local neighborhood information for node classification task.

\begin{table}
 \small
 \centering
    \begin{tabular}{l|cc|cc}
    \toprule[0.8pt]
    \multirow{2}{*}{\makecell*[c]{Transition \\ matrix}} & \multicolumn{2}{c|}{D $\rightarrow$ A}& \multicolumn{2}{c}{D $\rightarrow$ C}      \\ 
    \cmidrule[0.8pt]{2-5} 
                                      & Ma-F1 & Mi-F1 & Ma-F1 & Mi-F1  \\
    \cmidrule[0.8pt]{1-5} 
    \ \ \ \ $ \mathbf{P}_{\text {sym }}$      & 75.69 & 74.12 & 79.74 & 81.54  \\ 
    \ \ \ \ $ \mathbf{P}_{\text {no-loop }}$  & 75.96 & 74.51 & 78.69 & 80.39  \\ 
    \ \ \ \ $ \mathbf{P}_{\mathrm{rw}}$       & 75.77 & 74.04 & 78.10 & 79.72  \\
    \ \ \ \ $ \mathbf{P}_{\text {diff}}$      & 74.63 & 73.18 & 77.48 & 79.17  \\
    \bottomrule[0.8pt]
    \end{tabular}
    \caption{Performance with different transition matrices in the \propagate layer of \model.}
\label{tab:schemes}
\end{table}

\subsubsection{The effectiveness of different modules.}
Table \ref{tab:ablation} demonstrates that the addition of each module contributes to the final results without any performance degradation. 
The first line, $\mathcal{L}_{cls}$, only considers the source domain information and yields unsatisfactory performance, indicating that the model cannot be effectively applied to the target graph due to the existence of domain shifts. 
The second line incorporates the reduction of domain discrepancy by adding commonly used MMD constraint $\mathcal{L}_{mmd}$, highlighting the importance of reducing domain discrepancy. 
Next, we show how our model utilizes the inherent \generalizability of GNNs without adding additional trainable parameters.
The third line illustrates that performing \propagate for $k$ times enhances the \generalizability of GNN, which is consistent with Lemma 1. 
The fourth line demonstrates that the asymmetric architecture of our proposed \model can further boost its performance, which is in line with Lemma 2 and Lemma 3. 
In conclusion, our proposed \model achieves the best performance, which validates the effectiveness of combining all the aforementioned modules.

\begin{table}[]
\small
\centering
\hspace{-.2cm}
\begin{tabular}{l|cc|cc}
    \toprule[0.8pt]
    \multirow{2}{*}{Modules} & \multicolumn{2}{c|}{D $\rightarrow$ A}& \multicolumn{2}{c}{D $\rightarrow$ C}      \\ 
    \cmidrule[0.8pt]{2-5} 
                            & Ma-F1 & Mi-F1 & Ma-F1 & Mi-F1  \\
    \cmidrule[0.8pt]{1-5} 
      $ \mathcal{L}_{cls}$   & 58.50 & 61.11 & 66.78 & 70.43  \\ 
    \ + $\mathcal{L}_{mmd}$  & 67.09 & 65.85 & 72.96 & 74.83 \\ 
    \ + $k$ \textbf{P} layers  & 72.41 & 72.10 & 78.00 & 80.47  \\
    \ + asym arch & \textbf{75.69} & \textbf{74.12} & \textbf{79.74} & \textbf{81.54} \\
    \bottomrule[0.8pt]
    \end{tabular}
    \caption{Performance contribution of each part in \model.}
\label{tab:ablation}
\end{table}

\subsubsection{The sensitivity of the trade-off parameter $\alpha$.}\label{sec:ablation_alpha}
To assess the sensitivity of the trade-off parameter $\alpha$, we conduct experiments on $D \rightarrow A$ and $D \rightarrow C$ tasks, as depicted in \Fig \ref{fig:ablation}(b). The results indicate that \model consistently performs well across the range of $[0.0, 1.0]$, suggesting its robustness in the optimization process. When $\alpha = 0$, we solely use the source graph, and the overall loss function is reduced to $\mathcal{L} = \mathcal{L}_{cls}$. The drop in performance highlights the importance of minimizing domain discrepancy. Thus, it's necessary to minimize the distribution discrepancy. We also have provided more detailed ablation studies in Appendix.

\section{Conclusion}
In this paper, we study the problem of unsupervised graph domain adaptation on node classification task. We start by conducting a series of experiments to reveal the inherent \generalizability of GNNs. These observations provide important insights into designing a simple but effective model \model for UGDA. To take a step further, we derive the theoretical analysis of the error bound of multi-layer GNNs and show how \model tightens the error bound. Extensive experiments on real-world datasets demonstrate the effectiveness of our \model framework. In the future, we will extend our model to more complicated scenarios, such as source-free unsupervised graph domain adaptation and open-set graph domain adaptation.

\section{Acknowledgments}
This work is supported in part by the National Natural Science Foundation of China (Grant No.62372408, 62106221), Zhejiang Provincial Natural Science Foundation of China
(Grant No. LTGG23F030005), and Ningbo Natural Science
Foundation (Grant No. 2022J183).

\bibliography{reference}

\begin{thebibliography}{43}
\providecommand{\natexlab}[1]{#1}

\bibitem[{Arghal, Lei, and Bidokhti(2022)}]{Robust21}
Arghal, R.; Lei, E.; and Bidokhti, S.~S. 2022.
\newblock Robust graph neural networks via probabilistic lipschitz constraints.
\newblock In \emph{Learning for Dynamics and Control Conference}, 1073--1085.
  PMLR.

\bibitem[{Ben-David et~al.(2009)Ben-David, Blitzer, Crammer, Kulesza, Pereira,
  and Vaughan}]{BenDavid2009ATO}
Ben-David, S.; Blitzer, J.; Crammer, K.; Kulesza, A.; Pereira, F.~C.; and
  Vaughan, J.~W. 2009.
\newblock A theory of learning from different domains.
\newblock \emph{Machine Learning}, 79: 151--175.

\bibitem[{Ben-David et~al.(2006)Ben-David, Blitzer, Crammer, and
  Pereira}]{BenDavid2006}
Ben-David, S.; Blitzer, J.; Crammer, K.; and Pereira, F.~C. 2006.
\newblock Analysis of Representations for Domain Adaptation.
\newblock In \emph{NIPS}.

\bibitem[{Blitzer, McDonald, and Pereira(2006)}]{blitzer2006domain}
Blitzer, J.; McDonald, R.; and Pereira, F. 2006.
\newblock Domain adaptation with structural correspondence learning.
\newblock In \emph{Proceedings of the 2006 conference on empirical methods in
  natural language processing}, 120--128.

\bibitem[{Dai et~al.(2022)Dai, Wu, Xiao, Shen, and Wang}]{AdaGCN}
Dai, Q.; Wu, X.-M.; Xiao, J.; Shen, X.; and Wang, D. 2022.
\newblock Graph Transfer Learning via Adversarial Domain Adaptation with Graph
  Convolution.
\newblock \emph{TKDE}.

\bibitem[{Gama, Bruna, and Ribeiro(2019{\natexlab{a}})}]{Gama2019Stability}
Gama, F.; Bruna, J.; and Ribeiro, A. 2019{\natexlab{a}}.
\newblock Stability Properties of Graph Neural Networks.
\newblock \emph{IEEE Transactions on Signal Processing}, 68: 5680--5695.

\bibitem[{Gama, Bruna, and Ribeiro(2019{\natexlab{b}})}]{Stability19}
Gama, F.; Bruna, J.; and Ribeiro, A. 2019{\natexlab{b}}.
\newblock Stability Properties of Graph Neural Networks.
\newblock \emph{IEEE Transactions on Signal Processing}, 68: 5680--5695.

\bibitem[{Ganin and Lempitsky(2015)}]{Ganin2015UnsupervisedDA}
Ganin, Y.; and Lempitsky, V.~S. 2015.
\newblock Unsupervised Domain Adaptation by Backpropagation.
\newblock \emph{ArXiv}, abs/1409.7495.

\bibitem[{Ganin et~al.(2016)Ganin, Ustinova, Ajakan, Germain, Larochelle,
  Laviolette, Marchand, and Lempitsky}]{GRL}
Ganin, Y.; Ustinova, E.; Ajakan, H.; Germain, P.; Larochelle, H.; Laviolette,
  F.; Marchand, M.; and Lempitsky, V.~S. 2016.
\newblock Domain-Adversarial Training of Neural Networks.
\newblock In \emph{JMLR}.

\bibitem[{Goodfellow et~al.(2020)Goodfellow, Pouget-Abadie, Mirza, Xu,
  Warde-Farley, Ozair, Courville, and Bengio}]{goodfellow2020generative}
Goodfellow, I.; Pouget-Abadie, J.; Mirza, M.; Xu, B.; Warde-Farley, D.; Ozair,
  S.; Courville, A.; and Bengio, Y. 2020.
\newblock Generative adversarial networks.
\newblock \emph{Communications of the ACM}, 63(11): 139--144.

\bibitem[{Gretton et~al.(2012)Gretton, Borgwardt, Rasch, Sch{\"o}lkopf, and
  Smola}]{MMD}
Gretton, A.; Borgwardt, K.~M.; Rasch, M.~J.; Sch{\"o}lkopf, B.; and Smola, A.
  2012.
\newblock A Kernel Two-Sample Test.
\newblock \emph{JMLR}, 13: 723--773.

\bibitem[{Grover and Leskovec(2016)}]{node2vec}
Grover, A.; and Leskovec, J. 2016.
\newblock node2vec: Scalable Feature Learning for Networks.
\newblock \emph{KDD}.

\bibitem[{Hamilton, Ying, and
  Leskovec(2017{\natexlab{a}})}]{hamilton2017inductive}
Hamilton, W.; Ying, Z.; and Leskovec, J. 2017{\natexlab{a}}.
\newblock Inductive representation learning on large graphs.
\newblock \emph{Advances in neural information processing systems}, 30.

\bibitem[{Hamilton, Ying, and Leskovec(2017{\natexlab{b}})}]{GraphSAGE}
Hamilton, W.~L.; Ying, Z.; and Leskovec, J. 2017{\natexlab{b}}.
\newblock Inductive Representation Learning on Large Graphs.
\newblock In \emph{NIPS}.

\bibitem[{Hassani and Khasahmadi(2020)}]{hassani2020contrastive}
Hassani, K.; and Khasahmadi, A.~H. 2020.
\newblock Contrastive multi-view representation learning on graphs.
\newblock In \emph{International conference on machine learning}, 4116--4126.
  PMLR.

\bibitem[{Kipf and Welling(2017)}]{GCN}
Kipf, T.; and Welling, M. 2017.
\newblock Semi-Supervised Classification with Graph Convolutional Networks.

\bibitem[{Klicpera, Bojchevski, and G{\"u}nnemann(2018)}]{klicpera2018predict}
Klicpera, J.; Bojchevski, A.; and G{\"u}nnemann, S. 2018.
\newblock Predict then propagate: Graph neural networks meet personalized
  pagerank.
\newblock \emph{arXiv preprint arXiv:1810.05997}.

\bibitem[{Li et~al.(2021)Li, Zhou, Xu, Huang, Wang, Xiong, Huang, Dou, and
  Xiong}]{li2021structure}
Li, S.; Zhou, J.; Xu, T.; Huang, L.; Wang, F.; Xiong, H.; Huang, W.; Dou, D.;
  and Xiong, H. 2021.
\newblock Structure-aware interactive graph neural networks for the prediction
  of protein-ligand binding affinity.
\newblock In \emph{Proceedings of the 27th ACM SIGKDD Conference on Knowledge
  Discovery \& Data Mining}, 975--985.

\bibitem[{Liu et~al.(2023)Liu, Li, Feng, Tran, Zhao, Qiang, and Li}]{StruRW}
Liu, S.; Li, T.; Feng, Y.; Tran, N.; Zhao, H.; Qiang, Q.; and Li, P. 2023.
\newblock Structural Re-weighting Improves Graph Domain Adaptation.
\newblock In \emph{International Conference on Machine Learning}.

\bibitem[{Long et~al.(2015)Long, Cao, Wang, and Jordan}]{Long2015LearningTF}
Long, M.; Cao, Y.; Wang, J.; and Jordan, M.~I. 2015.
\newblock Learning Transferable Features with Deep Adaptation Networks.
\newblock In \emph{Proceedings of the 32nd International Conference on
  International Conference on Machine Learning - Volume 37}, ICML'15, 97–105.
  JMLR.org.

\bibitem[{Long et~al.(2017)Long, Cao, Wang, and Jordan}]{Long2017ConditionalAD}
Long, M.; Cao, Z.; Wang, J.; and Jordan, M.~I. 2017.
\newblock Conditional Adversarial Domain Adaptation.
\newblock In \emph{Neural Information Processing Systems}.

\bibitem[{Lu et~al.(2022)Lu, Gan, Jin, Fu, Wang, and Zhang}]{lu2022make}
Lu, B.; Gan, X.; Jin, H.; Fu, L.; Wang, X.; and Zhang, H. 2022.
\newblock Make more connections: Urban traffic flow forecasting with
  spatiotemporal adaptive gated graph convolution network.
\newblock \emph{ACM Transactions on Intelligent Systems and Technology (TIST)},
  1--25.

\bibitem[{Pan and Yang(2010)}]{10TransferLearningSurvey}
Pan, S.~J.; and Yang, Q. 2010.
\newblock A Survey on Transfer Learning.
\newblock \emph{TKDE}, 22: 1345--1359.

\bibitem[{Paszke et~al.(2019)Paszke, Gross, Massa, Lerer, Bradbury, Chanan,
  Killeen, Lin, Gimelshein, Antiga, Desmaison, K{\"o}pf, Yang, DeVito, Raison,
  Tejani, Chilamkurthy, Steiner, Fang, Bai, and Chintala}]{PyTorch}
Paszke, A.; Gross, S.; Massa, F.; Lerer, A.; Bradbury, J.; Chanan, G.; Killeen,
  T.; Lin, Z.; Gimelshein, N.; Antiga, L.; Desmaison, A.; K{\"o}pf, A.; Yang,
  E.; DeVito, Z.; Raison, M.; Tejani, A.; Chilamkurthy, S.; Steiner, B.; Fang,
  L.; Bai, J.; and Chintala, S. 2019.
\newblock PyTorch: An Imperative Style, High-Performance Deep Learning Library.
\newblock In \emph{NeurIPS}.

\bibitem[{Pei et~al.(2018)Pei, Cao, Long, and Wang}]{Pei2018MultiAdversarialDA}
Pei, Z.; Cao, Z.; Long, M.; and Wang, J. 2018.
\newblock Multi-Adversarial Domain Adaptation.
\newblock In \emph{AAAI Conference on Artificial Intelligence}.

\bibitem[{Perozzi, Al-Rfou, and Skiena(2014)}]{DeepWalk}
Perozzi, B.; Al-Rfou, R.; and Skiena, S. 2014.
\newblock DeepWalk: online learning of social representations.
\newblock \emph{KDD}.

\bibitem[{Shen et~al.(2020)Shen, Dai, Chung, Lu, and Choi}]{ACDNE}
Shen, X.; Dai, Q.; Chung, K. F.-L.; Lu, W.; and Choi, K.-S.~T. 2020.
\newblock Adversarial Deep Network Embedding for Cross-network Node
  Classification.
\newblock \emph{AAAI}.

\bibitem[{Shen et~al.(2021)Shen, Dai, Mao, lai Chung, and Choi}]{CDNE}
Shen, X.; Dai, Q.; Mao, S.; lai Chung, F.; and Choi, K.-S.~T. 2021.
\newblock Network Together: Node Classification via Cross-Network Deep Network
  Embedding.
\newblock \emph{IEEE Transactions on Neural Networks and Learning Systems}, 32:
  1935--1948.

\bibitem[{Tzeng et~al.(2017)Tzeng, Hoffman, Saenko, and
  Darrell}]{Tzeng2017AdversarialDD}
Tzeng, E.; Hoffman, J.; Saenko, K.; and Darrell, T. 2017.
\newblock Adversarial Discriminative Domain Adaptation.
\newblock \emph{CVPR}, 2962--2971.

\bibitem[{Velickovic et~al.(2018)Velickovic, Cucurull, Casanova, Romero,
  Lio’, and Bengio}]{GAT}
Velickovic, P.; Cucurull, G.; Casanova, A.; Romero, A.; Lio’, P.; and Bengio,
  Y. 2018.
\newblock Graph Attention Networks.

\bibitem[{Venkateswara et~al.(2017)Venkateswara, Eusebio, Chakraborty, and
  Panchanathan}]{venkateswara2017deep}
Venkateswara, H.; Eusebio, J.; Chakraborty, S.; and Panchanathan, S. 2017.
\newblock Deep hashing network for unsupervised domain adaptation.
\newblock In \emph{Proceedings of the IEEE conference on computer vision and
  pattern recognition}, 5018--5027.

\bibitem[{Wang and Deng(2018)}]{18DomainAdaptationSurvey}
Wang, M.; and Deng, W. 2018.
\newblock Deep Visual Domain Adaptation: A Survey.
\newblock \emph{Neurocomputing}, 312: 135--153.

\bibitem[{Wu et~al.(2019)Wu, Souza, Zhang, Fifty, Yu, and
  Weinberger}]{wu2019simplifying}
Wu, F.; Souza, A.; Zhang, T.; Fifty, C.; Yu, T.; and Weinberger, K. 2019.
\newblock Simplifying graph convolutional networks.
\newblock In \emph{International conference on machine learning}, 6861--6871.
  PMLR.

\bibitem[{Wu, He, and Ainsworth(2023)}]{GRADE}
Wu, J.; He, J.; and Ainsworth, E. 2023.
\newblock Non-iid transfer learning on graphs.
\newblock In \emph{Proceedings of the AAAI Conference on Artificial
  Intelligence}, 9, 10342--10350.

\bibitem[{Wu et~al.(2020)Wu, Pan, Zhou, Chang, and Zhu}]{UDAGCN}
Wu, M.; Pan, S.; Zhou, C.; Chang, X.; and Zhu, X. 2020.
\newblock Unsupervised Domain Adaptive Graph Convolutional Networks.
\newblock \emph{Proceedings of The Web Conference 2020}.

\bibitem[{Xia et~al.(2021)Xia, Li, Wu, and Li}]{xia2021deepis}
Xia, W.; Li, Y.; Wu, J.; and Li, S. 2021.
\newblock DeepIS: Susceptibility estimation on social networks.
\newblock In \emph{Proceedings of the 14th ACM International Conference on Web
  Search and Data Mining}, 761--769.

\bibitem[{Xu et~al.(2018)Xu, Hu, Leskovec, and Jegelka}]{xu2018powerful}
Xu, K.; Hu, W.; Leskovec, J.; and Jegelka, S. 2018.
\newblock How powerful are graph neural networks?

\bibitem[{Yang et~al.(2022)Yang, Wu, Wang, and Yan}]{yang2022graph}
Yang, C.; Wu, Q.; Wang, J.; and Yan, J. 2022.
\newblock Graph neural networks are inherently good generalizers: Insights by
  bridging gnns and mlps.

\bibitem[{You et~al.(2023)You, Chen, Wang, and Shen}]{SpecReg}
You, Y.; Chen, T.; Wang, Z.; and Shen, Y. 2023.
\newblock Graph Domain Adaptation via Theory-Grounded Spectral Regularization.
\newblock In \emph{The Eleventh International Conference on Learning
  Representations}.

\bibitem[{Zhang et~al.(2022)Zhang, Sheng, Yin, Jiang, Xia, Gao, Yang, and
  Cui}]{Degradation22}
Zhang, W.; Sheng, Z.; Yin, Z.; Jiang, Y.; Xia, Y.; Gao, J.; Yang, Z.; and Cui,
  B. 2022.
\newblock Model Degradation Hinders Deep Graph Neural Networks.
\newblock \emph{Proceedings of the 28th ACM SIGKDD Conference on Knowledge
  Discovery and Data Mining}.

\bibitem[{Zhang et~al.(2021)Zhang, Du, Xie, and Wang}]{ASN}
Zhang, X.; Du, Y.; Xie, R.; and Wang, C. 2021.
\newblock Adversarial Separation Network for Cross-Network Node Classification.
\newblock \emph{CIKM}.

\bibitem[{Zhang et~al.(2018)Zhang, Yang, Bu, Zhou, Yu, Zhang, Ester, and
  Wang}]{ANRL}
Zhang, Z.; Yang, H.; Bu, J.; Zhou, S.; Yu, P.; Zhang, J.; Ester, M.; and Wang,
  C. 2018.
\newblock ANRL: Attributed Network Representation Learning via Deep Neural
  Networks.
\newblock In \emph{IJCAI}.

\bibitem[{Zhu et~al.(2023)Zhu, Jiao, Ponomareva, Han, and
  Perozzi}]{zhu2023explaining}
Zhu, Q.; Jiao, Y.; Ponomareva, N.; Han, J.; and Perozzi, B. 2023.
\newblock Explaining and Adapting Graph Conditional Shift.
\newblock arXiv:2306.03256.

\end{thebibliography}

\clearpage
\onecolumn
\section{Appendix}

\subsubsection{Theorem 1.}
Suppose the feature extractor $f$ and the classifier $c$ is $K_f$-Lipschitz and $K_c$-Lipschitz continuous, where the Lipschitz norm $\|f\|_{\text {Lip }}=\max _{G_1, G_2} \frac{\left\|f\left(G_1\right)-f\left(G_2\right)\right\|_2}{\eta\left(G_1, G_2\right)}=K_f$ holds for some graph distance measure $\eta$. Let $\mathcal{H}:=\{h: \mathcal{G} \rightarrow \mathcal{Y}\}$ be the set of bounded real-valued functions, i.e. $h=c \circ f \in \mathcal{H}$, with probability at least $1-\delta$ the target risk $\epsilon_{\mathrm{T}}(g, \hat{g})$ is bounded as in the following inequality \cite{SpecReg}:
\begin{equation}
\begin{aligned}
    \epsilon_{\mathrm{T}}(h, \hat{h}) 
    \leq \ 
    & \hat{\epsilon}_{\mathrm{S}}(h, \hat{h}) 
      + 2 K_f K_c D\left(\mathbb{P}_{\mathrm{S}}(G), \mathbb{P}_{\mathrm{T}}(G)\right)\\
    & + \sqrt{\frac{4 d}{n^s} \log \left(\frac{e n^s}{d}\right) + \frac{1}{n^s} \log \left(\frac{1}{\delta}\right)} 
    + \omega,
\end{aligned} 
\label{therorem_1}
\end{equation}
where $\hat{\epsilon}_{\mathrm{S}}(h, \hat{h})$ and $\hat{\epsilon}_{\mathrm{T}}(h, \hat{h})$ represent the empirical source and target risks respectively. $\hat{h}$ denotes the labelling function. $\omega=\min _{\|c\|_{\operatorname{Lip}} \leq K_c,\|f\|_{\operatorname{Lip}} \leq K_f}\left\{\epsilon_{\mathrm{S}}(h, h^*)+\epsilon_{\mathrm{T}}(h, h^*)\right\}$ is the combined error of the ideal hypothesis $h^*$ that we expect to be small. $D(\cdot)$ indicates the distance metric, which formulates the source and target error functions in a Reproducing Kernel Hilbert Space.

Theorem 1 indicates that the generalization gap depends on the domain divergence $2 K_f K_c D\left(\mathbb{P}_{\mathrm{S}}(G), \mathbb{P}_{\mathrm{T}}(G)\right)$, where the Lipschitz constant $K_f$ is a conceptual property related to the model that needs to be instantiated \cite{SpecReg}. Therefore, one way to tighten the bound is to regularize the Lipschitz constant $K_f$ of the feature extractor $f$.

\subsection{Proof for Lemma 1}

We construct a GNN by composing a graph filter and nonlinear mapping that $f\left(G\right)
=r\left(\sigma\left(\mathcal{S}\left(A\right) X W\right)\right)$, where $r$ is the mean/sum/max readout function to pool node representations. $\mathcal{S}$ is the polynomial function that $\mathcal{S}\left(A\right)=\sum_{k=0}^{\infty} s_k A^k$. $W \in \mathcal{R}^{D \times D^{\prime}}$ denotes the learnable weight matrix. The pointwise nonlinearity holds as $|\sigma(b)-\sigma(a)| \leq|b-a|, \forall a, b \in \mathcal{R}$. Based on these notations, a $l$-layer GNN can be constructed as $f(G) = r \circ f^{(l)} \circ f^{(l-1)}\circ \cdots \circ f^{(1)}(G)$. The $l$-layer GNN Lipschitz constant $K_f$ is instantiated in the following lemma.

\subsubsection{Lemma 1.} 
Suppose that the edge perturbation is bounded that $\forall G_1, G_2 \in \mathcal{G}$, $EP = \left\|A_1-P^* A_2 P^{* \mathrm{~T}}\right\|_{\mathrm{F}} \leq \varepsilon$ with the optimal permutation $P^*$, and there exists an eigenvalue $\lambda^* \in \mathcal{R}$ to achieve the maximum $\left|\mathcal{S}\left(\lambda^*\right)\right|<\infty$. 
Assuming $\|X\|_{\mathrm{op}} \leq 1$ and $\|W\|_{\mathrm{op}} \leq 1$ ( $\|\cdot\|_{\mathrm{op}}$ stands for operator norm), we can then calculate the Lipschitz constant $l$-layer GNN as follows:
\begin{equation}
    \begin{aligned}
        K_f =\max \left\{
T_1^{(l)} + \sum_{i=1}^{l-1}{(\prod \limits_{j=i+1}^{l}T_2^{(j)})T_1^{(i)}}
, \prod \limits_{i=0}^l T_2^{(i)} \right\},
    \end{aligned}
    \notag
\end{equation}
where $T_1^{(l)} = K_\lambda^{(l)}\left(1+\tau \sqrt{n}\right)EP+\varepsilon \cdot \mathcal{O}\left(EP^2\right)$ 
and $T^{(l)}_2 = \left|\mathcal{S}^{(l)}\left(\lambda_T\right)\right|$. $\tau$ stands for the eigenvector misalignment that can be bounded. 
$n=n_S + n_T$ means the total number of nodes in source and target graph. 
$\mathcal{O}\left( \cdot \right)$ is the remainder term defined in \cite{Gama2019Stability}, 
and $K_\lambda$ is the spectral Lipschitz constant that $\forall \lambda_i, \lambda_j,\left|\mathcal{S}\left(\lambda_i\right)-\mathcal{S}\left(\lambda_j\right)\right| \leq K_\lambda\left|\lambda_i-\lambda_j\right|$.

\textit{Proof.} Denote the optimal permutation matrix for $G_1, G_2$ as $P^*$
, $T_{1,1} = \left(1+\tau \sqrt{n}\right)EP$
, $T_{1,2} = \mathcal{O}\left(EP^2\right)$
, $T_{1,0} = K_{\lambda} T_{1,1} + T_{1,2}$, and $T_{2,0}^{(l)}=\text{max}\left(\left|\mathcal{S}^{(l)}\left(\Lambda_2\right)\right|\right)$. 
We first prove the following inequality for a $l$-layer GNN, which compute the difference of the encoder outputs:
\begin{equation}
\hspace{-3.75cm}
\begin{aligned}
\left\|f\left(G_1\right)-f\left(G_2\right)\right\|_2
\leq 
 T_{1,0}^{(l)} + \sum_{i=1}^{l-1}{(\prod \limits_{j=i+1}^{l}T_{2,0}^{(j)})T_{1,0}^{(i)}} 
 + \prod \limits_{i=0}^l T_{2,0}^{(i)} \left\|X_1-P^* X_2\right\|_{\mathrm{F}}.
\label{l1_1}
\end{aligned}
\end{equation}

We use mathematical induction to prove this inequality. (\textit{i}) \textit{Base Case}: \Fom \ref{l1_1} holds true for $l = 1$ \cite{SpecReg}. (\textit{ii}) \textit{Inductive Hypothesis}: Assume that the theorem holds true for $l = k$, where $k$ is an integer greater than or equal to 1. (\textit{iii}) \textit{Inductive Step}: We aim to prove that the theorem also holds true for $l = k + 1$:
\begin{equation}
\hspace{-2cm}
\begin{aligned}
& \left\|f\left(G_1\right)-f\left(G_2\right)\right\|_2 \\
&= 
\left\|\left(r \circ f^{(k+1)}\right) \circ f^{(k)}\left(G_1\right)-\left(r \circ f^{(k+1)}\right) \circ f^{(k)}\left(G_2\right)\right\|_2 \\
&\stackrel{(a)}{\leq} 
K_\lambda^{(k+1)}T_{1,1} + T_{1,2} + T_{2,0}^{(k+1)} \left\|X_1^{(k)}-P^* X_2^{(k)}\right\|_{\mathrm{F}} \\
&\stackrel{(b)}{\leq} 
K_\lambda^{(k+1)}T_{1,1} + T_{1,2} + T_{2,0}^{(k+1)} \left( T_{1,0}^{(k)} + \sum_{i=1}^{k-1}{(\prod \limits_{j=i+1}^{k}T_{2,0}^{(j)})T_{1,0}^{(i)}} + \prod \limits_{i=0}^l T_{2,0}^{(i)} \left\|X_1-P^* X_2\right\|_{\mathrm{F}} \right) \\
&=  
T_{1,0}^{(k+1)} + \sum_{i=1}^{k}{(\prod \limits_{j=i+1}^{k+1}T_{2,0}^{(j)})T_{1,0}^{(i)}} + \prod \limits_{i=0}^{k+1} T_{2,0}^{(i)} \left\|X_1-P^* X_2\right\|_{\mathrm{F}},
\end{aligned}
\label{l1_2}
\end{equation}
where (a) has been proven in \cite{SpecReg} and (b) is the inductive hypothesis.
By proving the base case and the inductive step, we can conclude that according to mathematical induction, \Fom \ref{l1_2} holds true for all integers greater than or equal to the initial value.

Next, following the same spirit in \cite{SpecReg}, to calculate the Lipschitz constant $K_f$ , we assure the inequality:
\begin{equation}
\hspace{-0.5cm}
\begin{aligned}
& \left\|f\left(G_1\right)-f\left(G_2\right)\right\|_2 \\
& \leq 
T_{1,0}^{(l)} + \sum_{i=1}^{l-1}{(\prod \limits_{j=i+1}^{l}T_{2,0}^{(j)})T_{1,0}^{(i)}} + \prod \limits_{i=0}^{l} T_{2,0}^{(i)} \left\|X_1-P^* X_2\right\|_{\mathrm{F}} \\
& \stackrel{(a)}{=} 
K_\lambda^{(l)}T_{1,1} + T_{1,2} 
+ \sum_{i=1}^{l-1}{(\prod \limits_{j=i+1}^{l}T_{2,0}^{(j)})K_\lambda^{(i)}T_{1,1}} 
 + \sum_{i=1}^{l-1}{(\prod \limits_{j=i+1}^{l}T_{2,0}^{(j)})T_{1,2}} 
+ \prod \limits_{i=0}^{l} T_{2,0}^{(i)} \left\|X_1-P^* X_2\right\|_{\mathrm{F}} \\
& =
\left(K_\lambda^{(l)}
+ \sum_{i=1}^{l-1}{(\prod \limits_{j=i+1}^{l}T_{2,0}^{(j)})K_\lambda^{(i)}})\right) T_{1,1}
 + \left(1 + \sum_{i=1}^{l-1}{(\prod \limits_{j=i+1}^{l}T_{2,0}^{(j)})}\right)T_{1,2}
+ \prod \limits_{i=0}^{l} T_{2,0}^{(i)} \left\|X_1-P^* X_2\right\|_{\mathrm{F}} \\
& \leq 
K_f \eta\left(G_1, G_2\right),
\label{l1_3}
\end{aligned}
\end{equation}
where (a) is due to $T_{1,0}^{(l)} = K_{\lambda}^{(l)}T_{1,1}+T_{1,2}$.
Following \cite{SpecReg}, we define the commonly-used matching distance as $\eta\left(G_1, G_2\right)=\min _{P \in \Pi}\left\{\left\|X_1-P X_2\right\|_{\mathrm{F}}+\left\|A_1-P A_2 P^{\top}\right\|_{\mathrm{F}}\right\}$ \cite{Stability19, Robust21}. Then, above inequality is:
\begin{equation}
\begin{aligned}
\left(
\left(1+\tau \sqrt{n}\right) K_\lambda^{(l)}
+ \left(1+\tau \sqrt{n}\right) \sum_{i=1}^{l-1}{(\prod \limits_{j=i+1}^{l}T_2^{(j)})K_\lambda^{(i)}}) -K_f
\right) 
EP 
+ \left(1 + \sum_{i=1}^{l-1}{(\prod \limits_{j=i+1}^{l}T_2^{(j)})}\right)T_{1,2} & \\
+ \left(\prod \limits_{i=0}^{l} T_2^{(i)} - K_f \right)\left\|X_1-P^* X_2\right\|_{\mathrm{F}} &
 \leq 0,
\end{aligned}
\end{equation}
which is necessary for:
\begin{equation}
\begin{aligned}
 \left(\left(1+\tau \sqrt{n}\right) K_\lambda^{(l)}
+ \left(1+\tau \sqrt{n}\right) \sum_{i=1}^{l-1}{(\prod \limits_{j=i+1}^{l}T_2^{(j)})K_\lambda^{(i)}}) -K_f\right) EP 
 + \left(1 + \sum_{i=1}^{l-1}{(\prod \limits_{j=i+1}^{l}T_2^{(j)})}\right)T_{1,2} 
 & \leq 0, \\
\left(\prod \limits_{i=0}^{l} T_2^{(i)} - K_f \right)\left\|X_1-P^* X_2\right\|_{\mathrm{F}} 
& \leq 0,
\end{aligned}
\end{equation}
which is equivalent to:
\begin{equation}
\hspace{-4.5cm}
\begin{aligned}
K_f \geq  
&  K_\lambda^{(l)}T_{1,1} + \varepsilon \cdot T_{1,2} 
+ \sum_{i=1}^{l-1}{(\prod \limits_{j=i+1}^{l}T_{2,0}^{(j)})
\left( 
K_\lambda^{(i)}T_{1,1} 
+ \varepsilon \cdot T_{1,2}
\right)
}, \\ 
& = T_{1}^{(l)} + \sum_{i=1}^{l-1}{(\prod \limits_{j=i+1}^{l}T_2^{(j)})T_{1}^{(i)}} \\
K_f \geq  
& \prod \limits_{i=0}^l T_2^{(i)}
\end{aligned}
\end{equation}
Let $K_f$ takes the larger value between them, we complete the proof.

\subsection{Proof for Lemma 2} 
\subsubsection{Lemma 2.}
Following the setting in Lemma 1, we use a single linear \MLP in conjunction with $k$ \propagate layers in \model as $f_M^{\left( l \right)}\left(G\right)=r\left(\sigma\left(\mathcal{S}^{\left( l \right)}\left(A^k\right) X W\right)\right)$. We rephrase $T_2$  as $T_2(x)$, $x$ is the number of \propagate layers. For $\forall k > 1$, there exists: 
\begin{equation}
    \begin{aligned}
    \prod \limits_{i=0}^l T_2^{(i)}(k) < \prod \limits_{i=0}^l T_2^{(i)}(1), 
    \end{aligned}
    \notag
\end{equation}

\textit{Proof.} $T^{(l)}_2 = \left|\mathcal{S}^{(l)}\left(\lambda_T\right)\right|$ equals the spectral radius \cite{SpecReg}, which is the maximum absolute eigenvalue of the adjacency matrix $A$ and provides a measure of the smoothness of the graph. According to the consistency of matrix norm, the following inequality holds:
\begin{equation}
    \begin{aligned}
    0 < T_2^{(l)}(k) \leq \left\|A^p\right\|^{\frac{1}{p}} \leq\|A\| = T_2^{(l)}(1)
    \end{aligned}
\end{equation}
When $k=1$, the equal sign holds. Thus, for any $l>1$, $T_2^{(l)}(k) < T_2^{(l)}(1)$ holds, then we complete the proof. Intuitively, as the number of propagations increases, the signals on the similar nodes become more consistent, and the graph becomes smoother. This leads to a decrease in the spectral radius and the eigenvalues of the adjacency matrix.

\subsection{Proof for Lemma 3}

\subsubsection{Lemma 3.} 
Following the setting in Lemma 2, if the source graph structure does not participate in the training procedure, then the corresponding GNN degenerates into $f_L\left(G\right)
=r\left(\sigma\left( X W\right)\right)$. We rephrase $T_1^{(l)}$ as $T_1^{(l)}(x)$, and $x$ indicates the number of \propagate layers on source graph. For $\forall k > 1$, there exists: 
\begin{equation}
\begin{aligned}
 T_1^{(l)}(0) + \sum_{i=1}^{l-1}{(\prod \limits_{j=i+1}^{l}T_2^{(j)}(k))T_1^{(i)}(0)} 
 < T_1^{(l)}(1) + \sum_{i=1}^{l-1}{(\prod \limits_{j=i+1}^{l}T_2^{(j)}(1))T_1^{(i)}(1)}.
\end{aligned}
\notag
\end{equation}

\textit{Proof.} As the Lipschitz constant $K_f=\max _{G_1, G_2} \frac{\left\|f\left(G_1\right)-f\left(G_2\right)\right\|_2}{\eta\left(G_1, G_2\right)}$ and there exist data distribution shift between the source graph and target graph. Thus, we suppose the distance between nodes belonging to the same graph is smaller than the distance between different graphs. The following two inequality holds:
\begin{equation}
\begin{aligned}
\left\|f\left(G_{S,1}\right)-f\left(G_{T,2}\right)\right\|_2
> \left\|f\left(G_{S,1}\right)-f\left(G_{S,2}\right)\right\|_2, \\
\left\|f\left(G_{S,1}\right)-f\left(G_{T,2}\right)\right\|_2
> \left\|f\left(G_{T,1}\right)-f\left(G_{T,2}\right)\right\|_2,
\end{aligned}
\end{equation}
where $G_{S,i}$ and $G_{T,i}$ is the ego-graph from source and target graph respectively. Thus, $EP$ can be rephrased as $EP = \left\|A_{S,1}-P^* A_{T,2} P^{* \mathrm{~T}}\right\|_{\mathrm{F}}$. If the source graph structure does not participate in the training procedure, then the corresponding GNN degenerates into $f_L\left(G\right)
=r\left(\sigma\left( X W\right)\right) = r\left(\sigma\left(I X W\right)\right)$. In this case, we represent $EP$ as $EP' = \left\|I-P^* A_T P^{* \mathrm{~T}}\right\|_{\mathrm{F}}$. Then, the inequality holds:
\begin{equation}
\hspace{2cm}
\begin{aligned}
EP'^2 
&= \left\|I-P^* A_T P^{* \mathrm{~T}}\right\|_{\mathrm{F}}^2 \\
&< \left\|I-P^* A_T P^{* \mathrm{~T}}\right\|_{\mathrm{F}}^2
    + \left\|A_S-I\right\|_{\mathrm{F}}^2 
    + Re(<\left.A_S-I, I-P^* A_T P^{* \mathrm{~T}}>\right) \\
&= \left\|A_S-I+I-P^* A_T P^{* \mathrm{~T}}\right\|_{\mathrm{F}}^2 \\
&=\left\|A_S-P^* A_T P^{* \mathrm{~T}}\right\|_{\mathrm{F}}^2 = EP^2.
\end{aligned}
\end{equation}
For $EP'$ and $EP$ both greater than 0, we can derive  $EP' < EP$. Next, for $T_1^{(l)}(1) = K_\lambda^{(l)}\left(1+\tau \sqrt{n}\right)EP+\varepsilon \cdot \mathcal{O}\left(EP^2\right)$ and $T_1^{(l)}(0) = K_\lambda^{(l)}\left(1+\tau \sqrt{n}\right)EP'+\varepsilon \cdot \mathcal{O}\left(EP'^2\right)$, we can derive $T_1^{(l)}(1) > T_1^{(l)}(0)$. Finally, substitute $T_1^{(l)}(0)$ and $T_2^{(l)}(k)$ into \Fom \ref{lamma1}, the proof complete.

\subsection{Detialed of Minimize Domain Discrepancy}
There are two common approaches to reducing domain shift in domain adaptation. One approach involves explicitly measuring the discrepancy between the source and target graphs. This can be achieved using techniques such as maximum mean discrepancy (MMD) \cite{MMD}.
The other approach involves employing an adversarial objective with a domain discriminator in an implicit manner. 
By minimizing the domain classification loss while maximizing the classification accuracy, the main task network learns to produce domain-invariant features that can be used for classification in both the source and target domains.

\subsubsection{Maximum Mean Discrepancy Eestimation.} 
MMD quantifies the discrepancy between the two domains by comparing their mean embeddings in the reproducing kernel Hilbert space (RKHS). To calculate MMD, we first map the samples from each domain to their corresponding feature space using a kernel function. Then, we calculate the mean embeddings of the two distributions. The MMD is then defined as the distance between these mean embeddings, representing the discrepancy between the domains. The squared value of MMD is estimated as follows:
\begin{equation}
\begin{aligned}
\mathcal{L}_{mmd} & =\frac{1}{n_S^2} \sum_{i=1}^{n_S} \sum_{j=1}^{n_S} k_l\left(\phi_l\left(\boldsymbol{x}_i^s\right), \phi_l\left(\boldsymbol{x}_j^s\right)\right) \\
& +\frac{1}{n_T^2} \sum_{i=1}^{n_T} \sum_{j=1}^{n_T} k_l\left(\phi_l\left(\boldsymbol{x}_i^t\right), \phi_l\left(\boldsymbol{x}_j^t\right)\right) \\
& -\frac{2}{n_S n_T} \sum_{i=1}^{n_S} \sum_{j=1}^{n_T} k_l\left(\phi_l\left(\boldsymbol{x}_i^s\right), \phi_l\left(\boldsymbol{x}_j^t\right)\right),
\end{aligned}
\label{mmd-loss}
\end{equation}
where $n_S$ and $n_T$ is the number of nodes of source and target graph. And $k$ denotes the selected kernel. $\boldsymbol{x}_i^s$ and $\boldsymbol{x}_i^t$ denote the feature of source and target node respectively. By minimizing the MMD value, we aim to align the distributions of the source and target graphs to reduce the domain shift.

\begin{table}[t]
\small
\centering
\caption{Performance contribution of each part in ${\rm \model}_{adv}$}
\begin{tabular}{l|cc|cc}
    \toprule[0.8pt]
    \multirow{2}{*}{Modules} & \multicolumn{2}{c|}{D $\rightarrow$ A}& \multicolumn{2}{c}{D $\rightarrow$ C}      \\ 
    \cmidrule[0.8pt]{2-5} 
                            & Ma-F1 & Mi-F1 & Ma-F1 & Mi-F1  \\
    \cmidrule[0.8pt]{1-5} 
      $ \mathcal{L}_{cls}$   & 63.51 & 63.41 & 68.89 & 70.93  \\ 
    \ + $\mathcal{L}_{adv}$  & 64.19 & 64.13 & 69.59 & 72.01 \\ 
    \ + $k$ \propagate layers  & 66.95 & 68.09 & 75.56 & 79.20  \\
    \ + asymmetric architecture & \textbf{73.81} & \textbf{72.17} & \textbf{78.18} & \textbf{80.08} \\
    \bottomrule[0.8pt]
    \end{tabular}
\label{tab:ablation_adv}
\end{table}

\begin{table}[t]
    \caption{Standard deviations of \model on social network.}
    \small
    \centering
    \begin{tabular}{l|cc|cc}
    \toprule[0.8pt]
    \multirow{2}{*}{Models}	& \multicolumn{2}{c|}{DE$\rightarrow$EN} & \multicolumn{2}{c}{EN$\rightarrow$DE}	\\
    \cmidrule[0.8pt]{2-5}
        & Ma-F1 & Mi-F1 & Ma-F1 & Mi-F1 \\
    \cmidrule[0.8pt]{1-5}
    \multirow{2}{*}{$\textbf{\model}_{mmd}$}
        & 58.44 & 59.54 & 61.42 & 63.90 \\
        & $\pm$ 0.32 & $\pm$ 0.37 & $\pm$ 0.84 & $\pm$ 0.20 \\

    \cmidrule[0.8pt]{1-5}
    \multirow{2}{*}{$\textbf{\model}_{adv}$}
        & 59.16 & 59.94 & 62.03 & 65.11 \\
        & $\pm$ 0.21 & $\pm$ 0.20 & $\pm$ 0.20 & $\pm$ 0.24 \\
    \bottomrule[0.8pt]
    \end{tabular}
    \label{tab:deviations-social}
\end{table}

\subsubsection{Adversarial Training.} 
The goal of adversarial training, similar to MMD, is to learn domain-invariant representations without bias towards specific domains. By incorporating the domain discriminator into the training process, the encoder is encouraged to learn features that are indistinguishable between the source and target domains, aligning their distributions in the learned representation space. In the most routine procedure \cite{ACDNE, ASN, SpecReg} for implementing adversarial training, the domain classification loss is defined as follows:
\begin{equation}
    \mathcal{L}_{adv}=-\frac{1}{n} \sum_{v_i \in V}\left(1-d_i\right) \log \left(1-\hat{d}_i\right)+d_i \log \left(\hat{d}_i\right)
\end{equation}
where $d_i$ is the ground-truth domain label of $v_i$. And $\hat{d}_i$ represents the predicted domain probability of $v_i$. $n = n_S+n_T$ means the total number of nodes for the source graph and the target graph. $V = \left\{V^s \cup V^t\right\}$ is the collection of source and target nodes.

In the adversarial training framework, a Gradient Reversal Layer (GRL) \cite{GRL} is often inserted between the encoder and the domain discriminator to facilitate simultaneous updates during backpropagation. The GRL operates by reversing the sign of the gradients during backpropagation. Overall, the adversarial approach leverages the domain discriminator to implicitly align the node representations across domains, allowing the model to adapt and perform well in both the source and target domains.

\subsection{Hardware \& Software Environment} 
The experiments are performed on one Linux servers (CPU: Intel(R) Xeon(R) Platinum 8163 CPU @ 2.50GHz, Operation system: Ubuntu 18.04.5 LTS). For GPU resources, one NVIDIA Tesla V100 card with 32G are utilized The python libraries we use to implement our experiments are Python 3.8 and Pytorch 1.11.0. The maximum time for training one epoch is no more than five minute.

\subsection{More Experimental Results}
\begin{figure*}[h]
    \centering
    \subfigure[]{
    \includegraphics[width=0.28\linewidth]{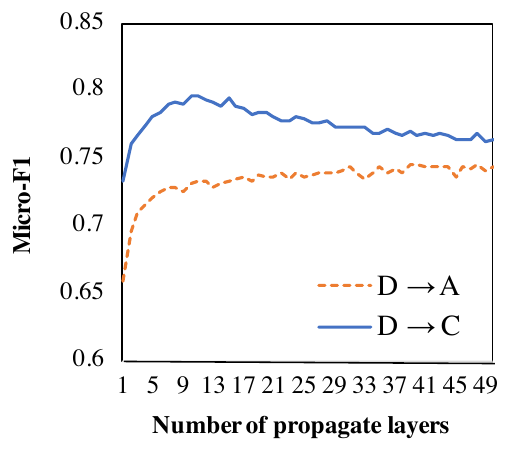}
    \label{fig:ablation-times-mif1-mmd}
    }
    \subfigure[]{
    \includegraphics[width=0.28\linewidth]{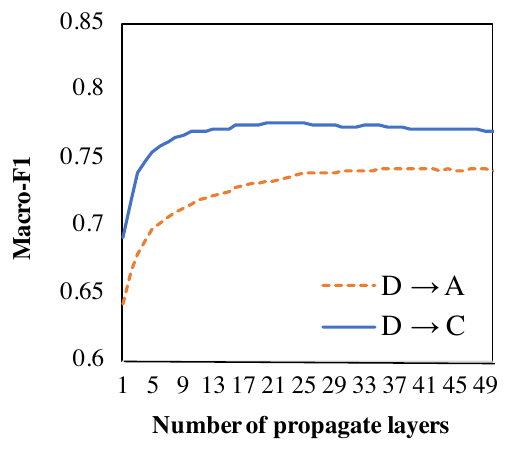}
    \label{fig:ablation-times-maf1-adv}
    }
    \subfigure[]{
    \includegraphics[width=0.28\linewidth]{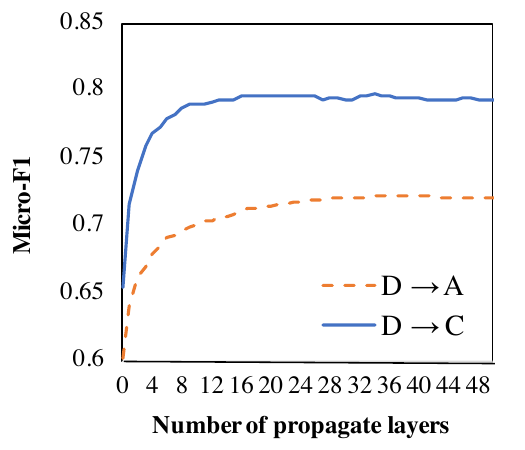}
    \label{fig:ablation-times-mif1-adv}
    }
    \subfigure[]{
    \includegraphics[width=0.28\linewidth]{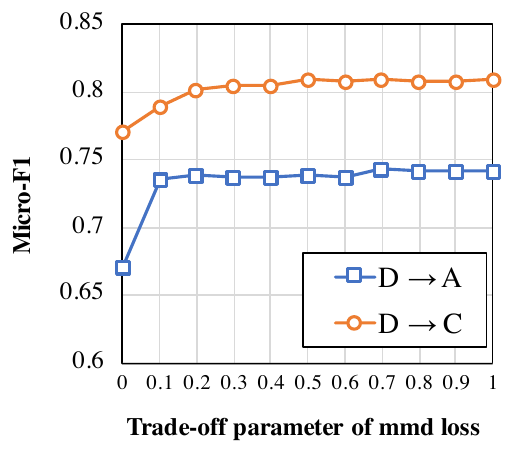}
    \label{fig:ablation-alpha-mif1-mmd}
    }
    \subfigure[]{
    \includegraphics[width=0.28\linewidth]{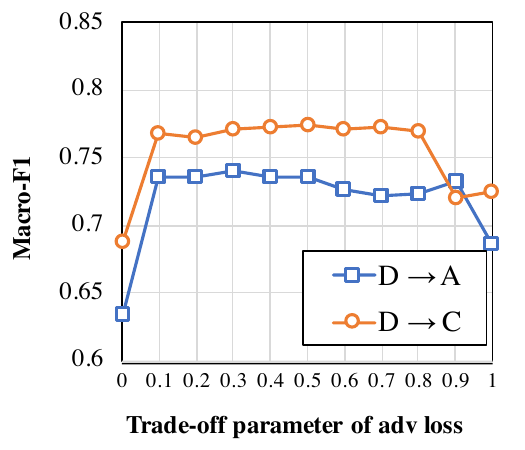}
    \label{fig:ablation-alpha-maf1-adv}
    }
    \subfigure[]{
    \includegraphics[width=0.28\linewidth]{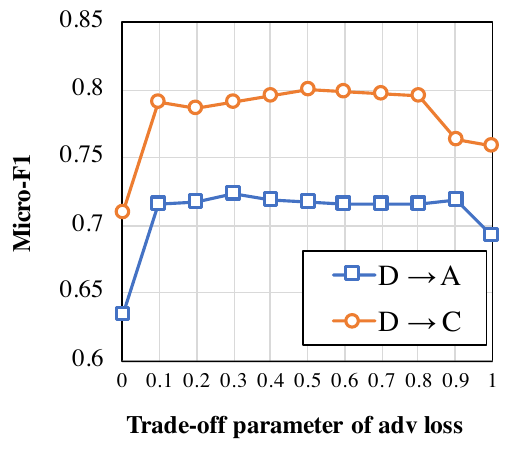}
    \label{fig:ablation-alpha-mif1-adv}
    }
    \caption{(a-c) shows the sensitivity of the number of \propagate layers $k$. (a) is the Micro-F1 of ${\rm \model}_{mmd}$. (b-c) is the Macro-F1 and Micro-F1 of ${\rm \model}_{adv}$ respectively. (d-f) shows the sensitivity of trade-off parameter $\alpha$. (d) is the Micro-F1 of ${\rm \model}_{mmd}$. (e-f) is the Macro-F1 and Micro-F1 of ${\rm \model}_{adv}$ respectively.}
    \label{fig:ablation_adv}
\end{figure*}

More experimental results are shown in \Fig \ref{fig:ablation_adv}. Among them, \Fig 
\ref{fig:ablation-times-mif1-mmd} and \ref{fig:ablation-alpha-mif1-mmd} is a supplement of ${\rm \model}_{mmd}$, which in line with the conclusion in Ablation Study. 
The remaining images validates the effectiveness of ${\rm \model}_{adv}$.  
Table \ref{tab:ablation_adv} demonstrates that the addition of each module in ${\rm \model}_{adv}$ contributes to the final results without any performance degradation. 
\Fig \ref{fig:ablation-times-maf1-adv} and \Fig \ref{fig:ablation-times-mif1-adv} demonstrate that utilizing a small number of $k \ (k<5)$ can significantly improve its performance, which is in line with the previous conclusion. For the $D \rightarrow A$ task, as the value of $k$ increases, its performance improves correspondingly. In contrast, for the $D \rightarrow C$ task, optimal performance is observed within the range of 5 to 10.
\Fig \ref{fig:ablation-alpha-maf1-adv} and \Fig \ref{fig:ablation-alpha-mif1-adv} demonstrate that ${\rm \model}_{adv}$ consistently performs well across the range of $[0.1, 0.8]$, suggesting its robustness in the optimization process. The optimal performance has a gradual decline as the value of $\alpha$ is greater than 0.8. The reason behind this is that, with an increase in the proportion of the adversarial loss function, the training of the model becomes gradually unstable, and the utilization of supervised signals weakens. 
The  standard deviations of the results are shown in \Tab \ref{tab:deviations-social}\&\ref{tab:deviations-citation}.

\begin{table*}[h]
    \centering
    \small 
    \caption{Standard deviations of \model on citation network.}
    \begin{tabular}{l|cc|cc|cc|cc|cc|cc}
    \toprule[0.8pt]
    \multirow{2}{*}{Models} 	&  \multicolumn{2}{c|}{D $\rightarrow$ A} & \multicolumn{2}{c|}{C $\rightarrow$ A} &\multicolumn{2}{c|}{A $\rightarrow$ D}	& \multicolumn{2}{c|}{C $\rightarrow$ D}	& \multicolumn{2}{c|}{A $\rightarrow$ C} & \multicolumn{2}{c}{D $\rightarrow$ C} \\
    \cmidrule[0.8pt]{2-13}
    & Ma-F1 & Mi-F1 & Ma-F1 & Mi-F1 & Ma-F1 & Mi-F1 & Ma-F1 & Mi-F1 & Ma-F1 & Mi-F1 & Ma-F1 & Mi-F1	\\
    \cmidrule[0.8pt]{1-13}
    \multirow{2}{*}{$\textbf{\model}_{mmd}$} 
        & 75.69 & 74.12 & 77.57 & 76.15 & 75.78 & 77.43 & 77.04 & 78.13 & 81.30 & 82.64 & 79.74 & 81.54 \\ 
        & $\pm$ 0.13 & $\pm$ 0.11 & $\pm$ 0.19 & $\pm$ 0.23 & $\pm$ 0.45 & $\pm$ 0.23 & $\pm$ 0.65 & $\pm$ 0.42 & $\pm$ 0.14 & $\pm$ 0.03 & $\pm$ 0.11 & $\pm$ 0.06 \\ 
    \cmidrule[0.8pt]{1-13}
    \multirow{2}{*}{$\textbf{\model}_{adv}$} 
        & 73.81 & 72.17 & 76.64 & 75.22 & 74.03 & 75.42 & 75.82 & 77.32 & 78.60 & 80.27 & 78.18 & 80.08 \\
        & $\pm$ 0.03 & $\pm$ 0.08 & $\pm$ 0.20 & $\pm$ 0.24 & $\pm$ 0.70 & $\pm$ 0.28 & $\pm$ 0.76 & $\pm$ 0.42 & $\pm$ 0.06 & $\pm$ 0.11 & $\pm$ 0.44 & $\pm$ 0.34 \\
    \bottomrule[0.8pt]
    \end{tabular}
    \label{tab:deviations-citation}
\end{table*}

\begin{table*}[!h]
    \caption{Symmetric architecture on social network.}
    \small
    \centering
    \begin{tabular}{l|cc|cc}
    \toprule[0.8pt]
    \multirow{2}{*}{Models}	& \multicolumn{2}{c|}{DE$\rightarrow$EN} & \multicolumn{2}{c}{EN$\rightarrow$DE}	\\
    \cmidrule[0.8pt]{2-5}
        & Ma-F1 & Mi-F1 & Ma-F1 & Mi-F1 \\
    \cmidrule[0.8pt]{1-5}
    graph-MLP & 54.14 & 57.10 & 54.23 & 55.74 \\
    $\textbf{\model}_{mmd}$ & 58.44 & 59.54 & 61.42 & 63.90 \\
    \bottomrule[0.8pt]
    \end{tabular}
    \label{tab:symmetric-social}
\end{table*}

\subsection{Explanation of different operations in Fig 1} 
Staking vanilla GCN layer denoted as [PT]. When stacking $n$ transformation operations, P[T] implies propagating on the graph before applying $n$ transformations with activation functions, while [T]P implies applying transformations before propagating once on the graph. For the implementation details, we use the vanilla GCNConv, and iterate linear transformation and propagation to achieve P[T] and [T]P respectively. Activation function is ReLU. For optimization, we utilize the Adam optimizer with a learning rate of 0.005 and a weight decay of 0.001.

\subsection{Implementation details of Tab. 1\&2}
The experimental settings of Tab. 1\&2 are consistent with the experimental section, which conducts node classification on citation dataset (D$\rightarrow$A) as described in Section: Empirical Analysis. Activation function is ReLU. For optimization, we utilize the Adam optimizer with a learning rate of 0.005 and a weight decay of 0.001. We set the trade-off parameter to 1. Our observations 1\&2 remain consistent across a range of datasets and tasks. The results of symmetric architecture with 1 transformation and k propagation layers are shown in Tab. \ref{tab:symmetric-citation} and \ref{tab:symmetric-social}.

\begin{table*}[t]
    \centering
    \small 
    \caption{Symmetric architecture on citation network.}
    \begin{tabular}{l|cc|cc|cc|cc|cc|cc}
    \toprule[0.8pt]
    \multirow{2}{*}{Models} 	&  \multicolumn{2}{c|}{D $\rightarrow$ A} & \multicolumn{2}{c|}{C $\rightarrow$ A} &\multicolumn{2}{c|}{A $\rightarrow$ D}	& \multicolumn{2}{c|}{C $\rightarrow$ D}	& \multicolumn{2}{c|}{A $\rightarrow$ C} & \multicolumn{2}{c}{D $\rightarrow$ C} \\
    \cmidrule[0.8pt]{2-13}
    & Ma-F1 & Mi-F1 & Ma-F1 & Mi-F1 & Ma-F1 & Mi-F1 & Ma-F1 & Mi-F1 & Ma-F1 & Mi-F1 & Ma-F1 & Mi-F1	\\
    \cmidrule[0.8pt]{1-13}
    graph-MLP
        & 73.56 & 71.82 & 76.09 & 75.03 & 72.93 & 77.13 & 74.54 & 77.69 & 80.73 & 82.48 & 77.85 & 80.39 \\ 
    $\textbf{\model}_{mmd}$
        & 75.69 & 74.12 & 77.57 & 76.15 & 75.78 & 77.43 & 77.04 & 78.13 & 81.30 & 82.64 & 79.74 & 81.54 \\ 
    \bottomrule[0.8pt]
    \end{tabular}
    \label{tab:symmetric-citation}
\end{table*}

\begin{table*}[t]
    \centering
    \small 
    \caption{Performance on source graphs. }
    \begin{tabular}{l|c|c|c|c|c|c|c|c}
    \toprule[0.8pt]
    Models 	&  D $\rightarrow$ A & {C $\rightarrow$ A} & {A $\rightarrow$ D}	& {C $\rightarrow$ D}	& {A $\rightarrow$ C} & {D $\rightarrow$ C} & DE$\rightarrow$EN & EN$\rightarrow$DE \\
    \midrule[0.8pt]
    GCN & 85.25 & 85.84 & 84.08 & 85.04 & 84.24 & 85.84 & 67.29 & 67.38 \\
    $\textbf{\model}_{mmd}$ & -0.04 & +0.02 & -0.62 & +0.40 & -0.79 & -0.15 & -0.67 & -1.20 \\ 
    $\textbf{\model}_{adv}$ & -0.06 & +0.02 & +0.08 & +0.94 & +1.38 & -0.02 & -3.45 & +0.68 \\
    \bottomrule[0.8pt]
    \end{tabular}
    \label{tab:source-acc}
\end{table*}

\subsection{Performance on source graph}
When transferring knowledge to the target graph, it is important to note that the decrease in the effectiveness of the source domain should not be considered as a cost. Therefore, we present the performance on the source graph with a baseline method called GCN, which is trained solely using the source graph. The experiments are repeated five times, and we report the mean performance in terms of Mi-F1. As can be seen in \Tab \ref{tab:source-acc}, the performance on source graphs  remains consistently high throughout the experiments.

\end{document}